\documentclass[journal]{IEEEtran}
\usepackage{ifpdf}
\usepackage{hyperref}
\hypersetup{
    colorlinks=true,
    urlcolor=blue,
    linkcolor=black}
\usepackage{algorithm2e}
\usepackage{array}
\usepackage{amsmath}
\usepackage{amssymb}  
\usepackage{booktabs}
\usepackage{caption}
\usepackage{color}
\usepackage{etoolbox}
\usepackage{epsfig}
\usepackage{filecontents}
\usepackage{graphicx}
\usepackage{graphics}
\usepackage{lmodern}
\usepackage{lipsum}
\usepackage{makecell}
\usepackage{siunitx}
\usepackage{soul}
\usepackage{slashbox}
\usepackage{subfig}
\usepackage{times}
\usepackage{textcomp}
\usepackage{verbatim}
\usepackage{xspace}
\usepackage{colortbl}
\usepackage{mathtools}
\usepackage{hyphenat}
\usepackage[table]{xcolor}

\graphicspath{{./figures/}}
\DeclareGraphicsExtensions{.png , .jpg, .jpeg, .eps}

\usepackage[noadjust]{cite}

\DeclareCaptionType{myalg}[{\bf Algorithm}]

\definecolor{maroon}{cmyk}{0,0.87,0.68,0.32}
\definecolor{Gray}{gray}{0.9}
\definecolor{LightCyan}{rgb}{0.88,1,1}
\definecolor{Red}{rgb}{1,0,0}
\definecolor{Green}{rgb}{0,1,0}
\definecolor{Blue}{rgb}{0,0,1}

\makeatletter
\DeclareRobustCommand\onedot{\futurelet\@let@token\@onedot}
\def\@onedot{\ifx\@let@token.\else.\null\fi\xspace}

\def\eg{\emph{e.g}\onedot} 
\def\ie{\emph{i.e}\onedot} 
 
\def\etc{\emph{etc}\onedot} \def\vs{\emph{vs}\onedot}
 
\def\etal{\emph{et al}\onedot}
\makeatother

\newcommand{\Fig}[1]{Fig. \ref{fig:#1}}
\newcommand{\Eq}[1]{Eq. (\ref{eq:#1})}
\newcommand{\Sect}[1]{Sect. \ref{sec:#1}}
\newcommand{\sSect}[1]{Sect. \ref{ssec:#1}}
\newcommand{\Tab}[1]{Table \ref{tab:#1}}
\newcommand{\Alg}[1]{Alg. \ref{alg:#1}}

\newcommand{\itn}[1]{\emph{#1}}

\makeatletter
\newcommand{\textString}[1]{#1\renewcommand{\@currentlabel}{#1}}
\makeatother

\ifCLASSINFOpdf
\else
\fi
\usepackage{amsmath}
\usepackage{algorithmic}
\newcommand{\R}{\mathbb{R}}

\newcommand{\realdata}{r}
\newcommand{\syntheticdata}{s}
\newcommand{\inputimage}{x}
\newcommand{\inputsequence}{X}
\newcommand{\groundtruth}{Y}
\newcommand{\depthmap}{d}
\newcommand{\semsegmap}{c}
\newcommand{\supmask}{\varpi}
\newcommand{\loss}{\mathcal{L}}
\newcommand{\supervised}{\mbox{{\scriptsize sp}}}
\newcommand{\lsupervised}{\mbox{{\scriptsize lsp}}}
\newcommand{\selfsupervised}{\mbox{{\scriptsize sf}}}
\newcommand{\domainadaptation}{\mbox{{\tiny DA}}}

\newcommand{\mde}{\Psi}
\newcommand{\wmde}{\theta}
\newcommand{\waux}{\vartheta}
\newcommand{\wauxself}{\waux^{\selfsupervised}}
\newcommand{\wauxda}{\waux^{\domainadaptation}}
\newcommand{\wmdeparenc}{\wmde^{\mbox{{\scriptsize enc}}}}
\newcommand{\wmdeparpyr}{\wmde^{\mbox{{\scriptsize pyr}}}}
\newcommand{\wmdepardec}{\wmde^{\mbox{{\scriptsize dec}}}}
\newcommand{\wmdeparpyrdec}{\wmde^{\mbox{{\scriptsize pyde}}}}
\newcommand{\lossequalizer}{\omega}
\newcommand{\maxestidepth}{d^{\mbox{{\scriptsize max}}}}

\newcommand{\meangtdepth}{d^{\syntheticdata,\mbox{{\scriptsize m}}}}
\newcommand{\meanestidepth}{\hat{d}^{\syntheticdata,\mbox{{\scriptsize m}}}}

\newcommand{\Nr}{N^{\realdata}}
\newcommand{\Ns}{N^{\syntheticdata}}
\newcommand{\ir}[1]{\inputimage^{\realdata}_{#1}} % e.g., $\ir{t}$ $\ir{t\pm1}$

\newcommand{\is}[1]{\inputimage^{\syntheticdata}_{#1}} % e.g., $\is{t}$ $\is{t\pm1}$
\newcommand{\dr}[1]{\depthmap^{\realdata}_{#1}} % e.g., $\dr{t}$
\newcommand{\ds}[1]{\depthmap^{\syntheticdata}_{#1}} % e.g., $\ds{t}$
\newcommand{\cs}[1]{\semsegmap^{\syntheticdata}_{#1}} % e.g., $\cs{t}$
\newcommand{\ms}[1]{\supmask^{\syntheticdata}_{#1}} % e.g., $\ms{t}$
\newcommand{\totalloss}{\loss}
\newcommand{\totallosspar}[1]{\totalloss(#1)}
\newcommand{\suploss}{\loss^{\supervised}}
\newcommand{\lsuploss}{\loss^{\lsupervised}}
\newcommand{\selfloss}{\loss^{\selfsupervised}}

\newcommand{\daloss}{\loss^{\domainadaptation}}
\newcommand{\suplosspar}[1]{{\suploss(#1)}}
\newcommand{\lsuplosspar}[1]{{\lsuploss(#1)}}
\newcommand{\selflosspar}[1]{{\selfloss(#1)}}

\newcommand{\dalosspar}[1]{{\daloss(#1)}}
\newcommand{\wmdepar}[1]{{\mde(\wmde;#1)}}
\newcommand{\RWSequences}{\inputsequence^{\realdata}}
\newcommand{\RWSequencesGT}{\groundtruth^{\realdata}}
\newcommand{\VWSequences}{\inputsequence^{\syntheticdata}}
\newcommand{\VWSequencesGT}{\groundtruth^{\syntheticdata}}
\newcommand{\selflosseq}{\lossequalizer^{\selfsupervised}}
\newcommand{\MiniBatch}{B}
\newcommand{\BRWSequences}{\inputsequence^{\realdata}_{\MiniBatch}}
\newcommand{\BRWSequencesGT}{\groundtruth^{\realdata}_{\MiniBatch}}
\newcommand{\BVWSequences}{\inputsequence^{\syntheticdata}_{\MiniBatch}}
\newcommand{\BVWSequencesGT}{\groundtruth^{\syntheticdata}_{\MiniBatch}}
\newcommand{\dalosseq}{\lossequalizer^{\domainadaptation}}
\newcommand{\LONEMETRIC}[1]{\|{#1}\|_1}
\newcommand{\lossgrad}[1]{{\nabla_{#1}}}
\newcommand{\lossgradvalue}[2]{{\Delta^{#1}_{#2}}}
\newcommand{\posenet}{\mathrm{T}}
\newcommand{\posetransf}{T}
\newcommand{\posenetpar}[1]{\posenet(#1)}
\newcommand{\posetransfpar}[1]{{\hat{\posetransf}}^{\realdata}_{#1}}
\newcommand{\irhpose}[1]{\hat{\inputimage}^{\realdata}_{#1}}
\newcommand{\cpe}{cpe}
\newcommand{\pe}{pe}
\newcommand{\cpepar}[1]{\cpe(#1)}
\newcommand{\pepar}[1]{\pe(#1)}
\newcommand{\averagedcpe}[1]{\overline{\cpe}(#1)}
\newcommand{\mr}[2]{\supmask^{\realdata}_{#1}(#2)}
\newcommand{\minphotoerror}[2]{P^{\realdata}_{#1}(#2)}
\newcommand{\mindepthedgeerror}[2]{S^{\realdata}_{#1}(#2)}

\newcommand{\danet}{\mathrm{D}}

\newcommand{\depthscalefactor}{\psi}

\newcommand{\gan}{\mathcal{G}}
\newcommand{\ganpar}[1]{{\gan(#1)}}
\newcommand{\wauxgan}{{\waux^{\gan}}}
\newcommand{\igan}{{\inputimage^{\gan}}}

\begin{document}
%\title{Monocular Depth Estimation through Real-world Structure-from-Motion Self-Supervision and Virtual-world Supervision}
\title{Monocular Depth Estimation through Virtual-world Supervision and Real-world SfM Self-Supervision}
% \begin{comment}

\author{Akhil Gurram$^{1,2}$,
        Ahmet Faruk Tuna$^{2}$,
        Fengyi Shen$^{2, 3}$,
        Onay Urfalioglu$^{2}$,
        and Antonio M. L\'opez$^{1,4}$
\thanks{$^{1}$Akhil and Antonio are with the Dpt. of Computer Science, Universitat Aut\`onoma de Barcelona (UAB), 08193 Bellaterra (Barcelona), Spain. $^{2}$Ahmet and Fengyi are with the Huawei European Research Center, 80992 M{\"u}nchen, Germany. Akhil and Onay were too during the development of this work. $^{3}$Fengyi is with the Dpt. of Informatics, Technische Universit\"at München (TUM), 85748, Garching, Germany. $^{4}$Antonio is also with the Computer Vision Center (CVC) at UAB, 08193 Bellaterra (Barcelona), Spain.}      
\thanks{Antonio acknowledges the financial support received for this research from the Spanish TIN2017-88709-R (MINECO/AEI/FEDER, UE) project. Antonio acknowledges the financial support to his general research activities given by ICREA under the ICREA Academia Program. Antonio acknowledges the support of the Generalitat de Catalunya CERCA Program as well as its ACCIO agency to CVC's general activities.}
}
\maketitle

\begin{abstract}
Depth information is essential for on-board perception in autonomous driving and driver assistance. Monocular depth estimation (MDE) is very appealing since it allows for appearance and depth being on direct pixelwise correspondence without further calibration. Best MDE models are based on Convolutional Neural Networks (CNNs) trained in a supervised manner, {\ie}, assuming pixelwise ground truth (GT). Usually, this GT is acquired at training time through a calibrated multi-modal suite of sensors. However, also using only a monocular system at training time is cheaper and more scalable. This is possible by relying on structure-from-motion (SfM) principles to generate self-supervision. Nevertheless, problems of camouflaged objects, visibility changes, static-camera intervals, textureless areas, and scale ambiguity, diminish the usefulness of such self-supervision. In this paper, we perform \emph{mono}cular \emph{d}epth \emph{e}stimation by \emph{v}irtual-world \emph{s}upervision (MonoDEVS) and real-world SfM self-supervision. We compensate the SfM self-supervision limitations by leveraging virtual-world images with accurate semantic and depth supervision, and addressing the virtual-to-real domain gap. Our MonoDEVSNet outperforms previous MDE CNNs trained on monocular and even stereo sequences.
\end{abstract}

% Note that keywords are not normally used for peer-review papers.

\section{Introduction}

Augmenting semantic information with depth is essential for on-board perception in autonomous driving and driver assistance. In this context, active sensors such as LiDAR and RADAR, or passive ones such as stereo rigs, are traditionally used to obtain depth information. For instance, in \cite{Dokhanchi:2021} RADAR and V2V communications are used to detect vehicles and estimate their distance to the ego-vehicle; LiDAR can be used for the same purpose \cite{Zhou:2018}, and it allows to perform road border detection too \cite{Deac:2019}; finally, note also that recent advances in deep stereo computation \cite{Cheng:2020} can bring stereo rigs as a LiDAR alternative for some driving use cases. However, due to cost and maintenance considerations, we wish to predict depth from the same single camera used to predict semantics, so having a direct pixelwise correspondence without further calibration. Therefore, in this paper, we focus on monocular depth estimation (MDE) on-board vehicles in outdoor traffic. Recent advances on MDE rely on Convolutional Neural Networks (CNNs). Let $\mde$ be a CNN architecture for MDE with weights $\wmde$, which takes a single image $\inputimage$ as input, and estimates its pixelwise depth map $\depthmap$ as output, {\ie}, $\wmdepar{\inputimage}\rightarrow\depthmap$. The $\mde\mbox{'s}$ can be trained in a supervised manner, {\ie}, finding the values of $\wmde$ by assuming access to a set of images with pixelwise depth ground truth (GT). Usually, such a GT is acquired at training time through a multi-modal suite of sensors, at least consisting of a camera calibrated with a LiDAR or some type of 3D laser scanner variant \cite{Eigen:2014, Liu:2016, Roy:2016, Laina:2016, Cao:2017, Fu:2018DORN, Gurram:2018, He:2018, Xu:2018, Yin:2019}. Alternatively, we can use self-supervision based on either a calibrated stereo rig \cite{Saxena:2007, Garg:2016, Godard:2017, Pillai:2019}, or a monocular system and structure-from-motion (SfM) principles \cite{Zhou:2017, Yin:2018GeoNet, Zhao:2020, Guizilini:20203D}, or a combination of both \cite{Godard:2019MonoDepth2}. Combining stereo self-supervision and LiDAR supervision has been also analyzed \cite{Kuznietsov:2017, He:2018wearable, Guizilini:2020}. The cheaper and simpler the suite of sensors used at training time, the better in terms of scalability and general access to the technology; however, the more challenging training a $\mde$. Currently, supervised methods tend to outperform self-supervised ones \cite{De:2021}, thus, improving the latter is an open challenge worth to pursue.   

We are interested in the most challenging setting, namely, when at training time we only have a single on-board camera allowing for SfM-based self-supervision. However, using only such a self-supervision may give rise to depth estimation inaccuracies due to camouflage (objects moving as the camera may not be distinguished from background), visibility changes (occlusion changes, non-Lambertian surfaces), static-camera cases ({\ie}, stopped ego-vehicle), and textureless areas, as well as to scale ambiguity (depth could only be estimated up to an unknown scale factor). In fact, an interesting approach to compensate for these problems could be leveraging virtual-world images (RGB) with accurate pixelwise depth (D) supervision. Using virtual worlds \cite{Gaidon:2016, Cabon:2020, Ros:2016, Mayer:2016, Richter:2017, Shah:2017, Dosovitskiy:2017}, we can acquire as many RGBD virtual-world samples as needed. However, these virtual-world samples can only be useful provided we address the virtual-to-real domain gap \cite{Zheng:2018T2Net, Kundu:2018AdaDepth, Zhao:2019GASDA, Pnvr:2020SharinGAN, Cheng:2020S3Net}, which links MDE with visual domain adaptation (DA), a realm of research in itself \cite{Csurka:2017, Wang-Deng:2018, Wilson:2020}. 

Accordingly, our contributions to MDE are the following:
\begin{itemize}
\item We propose a CNN architecture to perform MDE by training on virtual-world supervision and real-world SfM self-supervision, {\ie}, requiring just monocular sequences even at training time. We show that this architecture can accommodate different feature extraction backbones. 

\item We reduce domain discrepancies between supervised (virtual world) and semi-supervised (real world) data at the space of the extracted features (backbone bottleneck) by using the idea of gradient reversal layer (GRL) \cite{Ganin:2015, Ganin:2016}. Thus, not adding computational burden at testing time. 

\item Despite using SfM-based semi-supervision, thanks to the virtual-world supervised data, we can compute a global scaling factor at training time, which allows us to output absolute depth at testing time. 
\end{itemize}

In summary, we propose to perform \emph{mono}cular \emph{d}epth \emph{e}stimation by \emph{v}irtual-world \emph{s}upervision (MonoDEVS) and real-world SfM self-supervision, estimating depth in absolute scale. By relying on standard benchmarks, we show that our MonoDEVSNet outperforms previous ones trained on monocular and even stereo sequences. We think our released code and models\footnote{\href{https://github.com/HMRC-AEL/MonoDEVSNet}{https://github.com/HMRC-AEL/MonoDEVSNet}} will help researchers and practitioners to address applications requiring on-board depth estimation, also establishing a strong baseline to be challenged in the future.

In the following, Section \ref{sec:relatedwork} summarizes previous works related to ours. Section \ref{sec:method} details our proposal. Section \ref{sec:experiments} describes the experimental setting and discusses the obtained results. Finally, Section \ref{sec:conclusion} summarizes the presented work and conclusions, and draws the work we target for the near future. In addition, Appendix \ref{appsec:monodelsnet} introduces MonoDELSNet, where we replace virtual supervision by LiDAR supervision, showing also the corresponding MDE results.

\section{Related Work}
\label{sec:relatedwork}
MDE was first based on hand-crafted features and shallow machine learning \cite{Saxena:2007, Liu:2010, Ladicky:2014, Srikakulapu:2015}. Nowadays, best performing models are based on CNNs \cite{De:2021}, thus, we focus on them. 

\subsection{Supervised MDE}
\label{sec:rw:supervised}

Relying on depth GT, Eigen {\etal} \cite{Eigen:2014} developed a $\mde$ architecture for coarse-to-fine depth estimation with a scale-invariant loss function. This pioneering work inspired new CNN-based architectures to MDE \cite{Liu:2016, Laina:2016, Roy:2016, Cao:2017, He:2018, Xu:2018, Fu:2018DORN}, which also assume depth GT supervision. MDE has been also tackle as a task on a multi-task learning framework, typically together with semantic segmentation as both tasks aim at producing pixelwise information and, eventually, may help each other to improve their predictions at object boundaries. For instance, this is the case of some {$\mde$'s} for indoor scenarios \cite{Mousavian:2016, Jafari:2017, Jiao:2018}. These proposals assume that pixelwise depth and class GT are simultaneously available at training time. However, this is expensive, being scarcely available for outdoor scenarios. In order to address this problem, Gurram {\etal} \cite{Gurram:2018} proposed a training framework which does not require depth and class GT to be available for the same images. Guizilini {\etal} \cite{Guizilini:2020semantic} used an out-of-the-box CNN for semantic segmentation to train semantically-guided depth features while training $\mde$. 

The drawback of these supervised approaches is that the depth GT usually comes from expensive LiDARs, which must be calibrated and synchronized with the cameras; {\ie}, even if the objective is to use only cameras for the functionality under development. Moreover, LiDAR depth is sparse compared to available image resolutions. Besides, surfaces like vehicle glasses or dark vehicles may be problematic for LiDAR sensing. Consequently, depth self-supervision and alternative sources of supervision are receiving increasing interest. 

\subsection{Self-supervised MDE}
\label{sec:rw:self-supervised}

Using a calibrated stereo rig to provide self-supervision for MDE is a much cheaper alternative to camera-LiDAR suites. Garg {\etal} \cite{Garg:2016} pioneered this approach by training $\mde$ with a warping loss involving pairs of stereo images. Godard {\etal} \cite{Godard:2017} introduced epipolar geometry constraints with additional terms for smoothing and enforcing consistency between left-right image pairs. Chen {\etal} \cite{Chen:2019} improved MDE results by enforcing semantic consistency between stereo pairs, via a joint training of $\mde$ and semantic segmentation. Pillai {\etal} \cite{Pillai:2019} implemented sub-pixel convolutional layers for depth super-resolution, as well as a novel differentiable layer to improve depth prediction on image boundaries, a known limitation of stereo self-supervision. Other authors \cite{Kuznietsov:2017, He:2018wearable} still complement stereo self-supervision with sparse LiDAR supervision.

SfM principles \cite{Ozyesil:2017} can be also followed to provide self-supervision for MDE. In fact, in this setting we can assume a monocular on-board system at training time. Briefly, the underlying idea is that obtaining a frame, $\inputimage_{t}$, from consecutive ones, $\inputimage_{t\pm1}$, can be decomposed into jointly estimating the scene depth for $\inputimage_{t}$ and the camera pose at time $t$ relative to its pose at time $t\pm1$; {\ie}, the camera ego-motion. %, provided it does not vanish. 
Thus, we can train a CNN to estimate (synthesize) $\inputimage_{t}$ from $\inputimage_{t\pm1}$, where, basically, the photo-metric error between $\inputimage_{t}$ and $\hat{\inputimage}_{t}$ acts as training loss, being $\hat{\inputimage}_{t}$ the output of this CNN ({\ie}, the synthesized view). After the training process, part of the CNN can perform MDE up to a scale factor (relative depth).

Zhou {\etal} \cite{Zhou:2017} followed this idea, adding an explainability mask to compensate for violations of SfM assumptions (due to frame-to-frame changes on the visibility of frame's content, textureless surfaces, {\etc}). This mask is estimated by a CNN jointly trained with $\mde$ to output a pixelwise belief on the synthesized views. Later, Yin {\etal} \cite{Yin:2018GeoNet} proposed GeoNet, which aims at improving MDE by also predicting optical flow to explicitly consider the motion introduced by dynamic objects ({\eg}, vehicles, pedestrians), {\ie} a motion that violates SfM assumptions. However, this was effective on predicting occlusions, but not in significantly improving MDE accuracy. Godard {\etal} \cite{Godard:2019MonoDepth2} followed the idea of having a mask %too, in this case 
to indicate stationary pixels, which should not be taken into account by the loss driving the training. Such pixels typically appear on vehicles moving at the same speed as the camera, or can even correspond to full frames in case the ego-vehicle stops and, thus, the camera becomes stationary for a while. Pixels of similar appearance in consecutive frames are considered as stationary. A simple definition which can work because, instead of using a training loss based on absolute photo-metric errors ({\ie} on minimizing pairwise pixel differences), it is used the structure similarity index measurement (SSIM) \cite{Wang:2004}. Moreover, within the so-called  MonoDepth2 framework, Godard {\etal} \cite{Godard:2019MonoDepth2} combine SfM and stereo self-supervision to establish state-of-the-art results. Alternatively, Guizilini {\etal} \cite{Guizilini:2020semantic} addressed the presence of dynamic objects by a two-stage MDE training process. The first stage ignores the presence of such objects, returning a $\mde$ trained with a loss based on SSIM. Then, before running the second stage, the training sequences are processed to filter out frames that may contain erroneous depth estimations due to moving objects. Such frames are identified by applying $\mde$, a RANSAC algorithm to estimate the ground plane from their estimated depth, and determining if there is a significant number of pixels that would be projected far below the ground plane. Finally, in the second stage, $\mde$ is retrained form scratch without the filtered frames. 

Zhao {\etal} \cite{Zhao:2020} focused on avoiding scale inconsistencies among frames as produced by SfM self-supervision, specially when they are from sequences whose underlying depth range is too different. Depth and optical flow estimation CNNs are trained, but not a pose estimation one. Instead, the optical flow between two frames is used to find robust pixel correspondences between them, which are used to compute their relative camera pose, computing the fundamental matrix by the 8-point algorithm, and then performing triangulation between the corresponding pixels of these frames. Overall, a sparse depth pseudo-GT is estimated and used as supervision to train  $\mde$. However, even robustifying scale consistency among frames, this method still outputs just relative depth. To avoid this problem, Guizilini {\etal} \cite{Guizilini:2020} used sparse LiDAR supervision with SfM self-supervision, relying on depth and pose estimation networks. More recently, Guizilini {\etal} \cite{Guizilini:20203D} relied on ego-vehicle velocity to solve scale ambiguity in a pure SfM self-supervision setting. A velocity supervision loss trains the pose estimation CNN to learn scale-aware camera translation which, in turn, enables scale-aware depth estimation. 

Overall, this literature shows the relevance of achieving MDE via SfM self-supervision and strategies to account for violation of SfM assumptions, as well as to obtain absolute depth values. Among these strategies, complementing SfM self-supervision with supervision (depth GT) coming from additional sensors such as a LiDAR and/or a stereo rig seems to be the most robust approach to address all the problems at once. However, then, a single camera would not be enough at training time. In this paper, we also complement SfM self-supervision with accurate depth supervision. However, instead of relying on additional sensors, we use virtual-world data.

\subsection{Virtual-world data for MDE}
\label{sec:rw:virtual-world-MDE}

Training $\mde$ on virtual-world images to later perform on real-world ones, requires to address the virtual-to-real domain gap. Many approaches perform a virtual-to-real image-to-image translation coupled to the training of $\mde$. This translation usually relies on generative adversarial networks (GANs) \cite{Goodfellow:2014, Choi:2020}, since to train them only unpaired and unlabeled sets of real- and virtual-world images are required.

Zheng {\etal} \cite{Zheng:2018T2Net} proposed $T^2$Net. In this case, a GAN and $\mde$ are jointly trained, where the GAN aims at performing virtual-to-real translation while acting as an auto-encoder for real-world images. The translated images are the input for $\mde$ since they have depth supervision. Additionally, a GAN operating on the encoder weights (features) of $\mde$ was incorporated during training to force similar depth feature distributions between translated and real-world images. However, this feature-level GAN worsen MDE results in outdoor scenarios. Kundu {\etal} \cite{Kundu:2018AdaDepth} proposed AdaDepth, which trains a common feature space for real- and virtual-world images, {\ie}, a space where it is not possible to distinguish the domain of the input images. Then, depth estimation is trained from this feature space. To achieve this, adversarial losses are used at the feature space level as well as at the estimated depth level. 

Cheng {\etal} \cite{Cheng:2020S3Net} proposed $S^3$Net, which extends $T^2$Net with SfM self-supervision. In this case, GAN training involves semantic and photo-metric consistency. Semantic consistency between the virtual-world images and their GAN-translated counterparts is required, which is measured via semantic segmentation (which involves also to jointly train a CNN for this task). Photo-metric consistency is required for consecutive GAN-translated images, which is measured via optical flow. Note that semantic segmentation and optical flow GT is available for virtual-world images. $\mde$ uses the GAN-translated images as input and is trained end-to-end with the GAN. Then, a further fine-tuning step of $\mde$ is performed using only the real-world sequence and SfM self-supervision, {\ie}, involving the training of a pose estimation CNN while fine-tuning. During this process, a masking mechanism inspired in \cite{Godard:2019MonoDepth2} is also used to compensate for SfM-adverse scenarios. Contrary to AdaDepth and $T^2$Net, $S^3$Net just outputs relative depth. 

Zhao {\etal} \cite{Zhao:2019GASDA} proposed GASDA, which leverages real-world stereo and virtual-world data. In this case, the CycleGAN idea \cite{Zhu:2017} is used to perform DA, which actually involves two GANs, one for virtual-to-real image translation and another for real-to-virtual. Two $\mde\mbox{'s}$ are trained coupled to CycleGAN, one intended to process images with real-world appearance (actual real-wold images or GAN-translated from the virtual domain), the other to process images with synthetic appearance (actual virtual-world images or GAN-translated from the real domain). In fact, at testing time, the most accurate depth results are obtained by averaging the output of these two $\mde\mbox{'s}$, which also involves to translate the real-world images to the virtual domain by the corresponding GAN. Thanks to the stereo data, left-right depth and geometry consistency losses are also included during training aiming at obtaining a more accurate $\mde$. PNVR {\etal} \cite{Pnvr:2020SharinGAN} proposed SharinGAN for training a DA GAN coupled to a specific task. One of the selected tasks is MDE with stereo self-supervision, as in \cite{Zhao:2019GASDA}. In this case, real- and virtual-world images are transformed to a new image domain where their appearance discrepancies are minimized to perform MDE from them, {\ie} the GAN and the $\mde$  are jointly trained end-to-end. SharinGAN outperformed GASDA. However, at testing time, before performing the MDE, the real-world images must be translated by the GAN to the new image domain. Both GASDA and SharinGAN produce absolute scale depth.

\subsection{Relationship of MonoDEVSNet with previous literature} 
In term of operational training conditions, the most similar paper to ours is $S^3$Net \cite{Cheng:2020S3Net}. However, contrary to $S^3$Net, our MonoDEVSNet can estimate depth in absolute scale. On the other hand, for the SfM self-supervision we leverage from the state-of-the-art proposal in \cite{Godard:2019MonoDepth2}. Note that methods based on pure SfM self-supervision such as \cite{Godard:2019MonoDepth2} (only SfM setting), \cite{Zhou:2017}, \cite{Yin:2018GeoNet}, and \cite{Guizilini:2020semantic}, just report relative depth. In order to compare MonoDEVSNet with them, we have estimated relative depth too. We will see how we outperform these methods, proving the usefulness of leveraging depth supervision from virtual worlds. In fact, regarding relative depth, we also outperform $S^3$Net. Methods leveraging virtual-world data such as GASDA \cite{Zhao:2019GASDA} and SharinGAN \cite{Pnvr:2020SharinGAN}, rely on real-world stereo data at training time, while we only require monocular sequences. On the other hand, our training framework can be extended to accommodate stereo data if available, although it is not our current focus. $S^3$Net, GASDA, SharinGAN, $T^2$Net \cite{Zheng:2018T2Net}, and AdaDepth \cite{Kundu:2018AdaDepth}, leverage ideas from GAN-based DA to reduce the virtual-to-real domain gap, either in image space ($S^3$Net, GASDA, SharinGAN, $T^2$Net) or in feature space (AdaDepth). We have analyzed both, image and feature based DA, finding that the later outperforms the former. In particular, by using the Gradient-Reversal-Layer (GRL) DA strategy \cite{Ganin:2015, Ganin:2016}, up to the best of our knowledge, not previously applied to MDE. Currently, we outperform the SfM self-supervision framework in \cite{Guizilini:20203D} thanks to the virtual-world supervision and our GRL DA strategy. However, using vehicle velocity to obtain absolute depth as in \cite{Guizilini:20203D}, is a complementary strategy that could be also incorporated in our framework, although it is not the focus on this paper.  

\section{Methods}
\label{sec:method}
\begin{figure}
    \centering
    \includegraphics[clip=True, trim=0 0 0 0,width=\columnwidth]{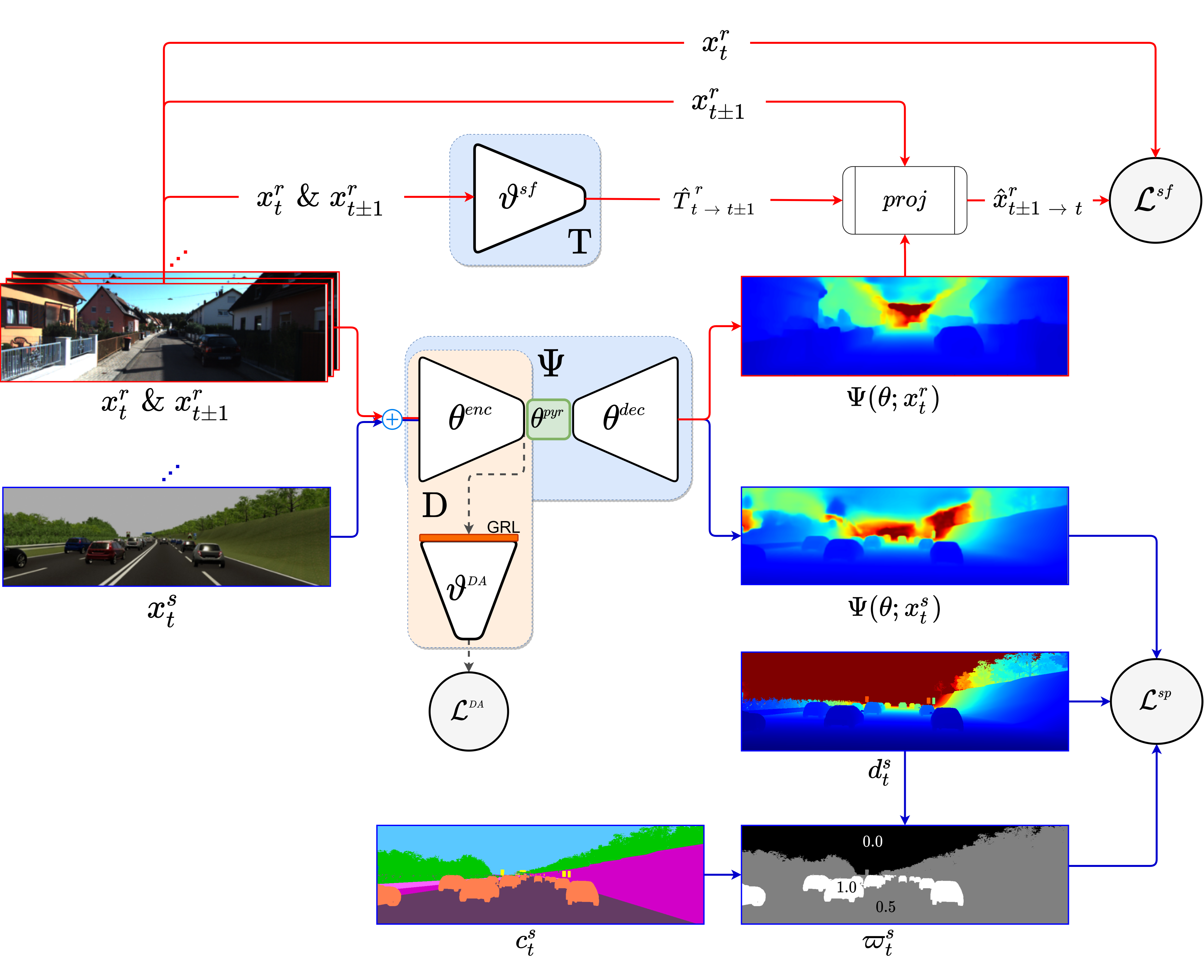}
    \caption{Training framework for MonoDEVSNet, {\ie}, $\wmdepar{\inputimage}$. We show the involved images, GT, weights, and losses. Red and blue lines are paths of real and virtual-world data, respectively. The discontinuous line is a common path. 
    }
    \label{fig:overall_training}
\end{figure}

In this section, we introduce MonoDEVSNet, which aims at leveraging virtual-world supervision to improve real-world SfM self-supervision. Since we train from both real- and virtual-world data jointly, we describe our supervision and self-supervision losses, the loss for addressing the virtual-to-real domain gap, and the strategy to obtain depth in absolute scale. Our proposal is visually summarized in \Fig{overall_training}.

\subsection{Training data}
\label{ssec:trainingdata}
For training MonoDEVSNet, we assume two sources of  data. On the one hand, we have image sequences acquired by a monocular system on-board a vehicle while driving in real-world traffic. We denote as $\ir{t}$ one of such frames acquired at time $t$. We denote these data as $\RWSequences=\{\ir{t}\}_{t=1}^{\Nr}$, where $\Nr$ is the number of frames from the real-world sequences. These frames do not have associated GT. On the other hand, we have analogous sequences but acquired on a virtual world, {\ie}, on-board a vehicle immersed in a traffic simulation. We denote as $\is{t}$ one of such virtual-world frames acquired at time $t$. We refer to these data as $\VWSequences=\{\is{t}\}_{t=1}^{\Ns}$, where $\Ns$ is the number of frames from the virtual-world sequences. The images in $\VWSequences$ do have associated GT, since it can be automatically generated. In particular, as it is commonly available in today's simulators, we assume pixelwise depth and semantic class GT. We define $\VWSequencesGT=\{<\ds{t},\cs{t}>\}_{t=1}^{\Ns}$ to be this GT; {\ie}, given $\is{t}$, $\ds{t}$ is its depth GT, and $\cs{t}$ its semantic class GT. 

\subsection{MonoDEVSNet architecture: $\wmdepar{\inputimage}$}
\label{ssec:MDECNN}

MonoDEVSNet, {\ie}, our $\wmdepar{\inputimage}$, is composed of three main blocks: a encoding block of weights $\wmdeparenc$, a multi-scale pyramidal block, $\wmdeparpyr$, and a decoding block inspired in \cite{Godard:2019MonoDepth2}, $\wmdepardec$. Therefore, the total set of weights is $\wmde=\{\wmdeparenc,\wmdeparpyr,\wmdepardec\}$. Here, $\wmdeparenc$ acts as a backbone of features. Moreover, since we aim at evaluating several encoders, the role of the multi-scale pyramid block is to adapt the bottleneck of the chosen encoder to the decoder. At testing time $\wmdepar{\inputimage}$ will process any real-world image $\inputimage$ acquired on-board the ego-vehicle, while at training time either $\inputimage\in\RWSequences$ or $\inputimage\in\VWSequences$.

\subsection{Problem formulation}
\label{ssec:formulation}

Training $\wmdepar{\inputimage}$ consists in finding the optimum weight values, $\wmde^*$, by solving the problem:
\begin{equation}
\displaystyle{\wmde^* = \min_{\wmde} \totallosspar{\wmde;\RWSequences,{\VWSequences.\VWSequencesGT}}} \enspace , \notag  
\label{eq:formulation}
\end{equation}
\noindent where $\totalloss$ is a loss function, and ${\VWSequences.\VWSequencesGT}$ indicates the use of the virtual-world frames with their GT. As we are going to detail, $\totalloss$ relies on three different losses, namely, $\selfloss, \suploss$ and $\daloss$. The loss $\selfloss$ focuses on training $\wmde$ based on SfM self-supervision, thus, only relying on real-world data sequences. The SfM self-supervision is achieved with the support of a camera pose estimation task performed by a CNN, $\posenet$, of weights $\wauxself$. Thus, we have $\selflosspar{\wmde, \wauxself; {\RWSequences}}$. The loss $\suploss$ focuses on training $\wmde$ with virtual-world supervision, in particular, using depth and semantic GT from virtual-world sequences. Therefore, we have $\suplosspar{\wmde; {\VWSequences.\VWSequencesGT}}$. Finally, $\daloss$ focuses on creating domain-invariant features $\wmdeparenc$ as part of $\wmde$. In particular, we rely on a binary real/virtual domain-classifier CNN, $\danet$, of weights $\{\wmdeparenc,\wauxda\}$. Thus, we have $\dalosspar{\wmdeparenc, \wauxda; \RWSequences, \VWSequences}$.

\subsection{SfM Self-supervised loss: $\selflosspar{\wmde, \wauxself; {\RWSequences}}$}
\label{ssec:slfloss}
Since we focus on improving MDE by the additional use of virtual-world data, for the SfM self-supervision we leverage from the state-of-the-art proposal in \cite{Godard:2019MonoDepth2}, which we briefly summarize here for the sake of completeness as:
\begin{align}
\label{eq:sf-loss}
&\selflosspar{\wmde, \wauxself; \RWSequences} = \sum_{t=2}^{\Nr-1} \minphotoerror{t}{\wmde,\wauxself} + \lambda \mindepthedgeerror{t}{\wmde} \enspace .
\end{align}
As introduced in \cite{Godard:2017}, the term $\lambda \mindepthedgeerror{t}{\wmde}$ is a constant weighted loss to force local smoothness on $\wmdepar{\ir{t}}$, taking into account the edges of $\ir{t}$. The term $\minphotoerror{t}{\wmde,\wauxself}$ is the actual SfM-inspired loss. It involves the joint training of the depth estimation weights, $\wmde$, and the relative camera pose estimation weights, $\wauxself$. Figure \ref{fig:overall_training} illustrates the CNN, $\posenet$, associated to these weights, which takes as input two consecutive frames, {\eg}, $(\ir{t},\ir{t+1})$, and outputs the pose transform (rotation and translation), $\posetransfpar{t\rightarrow t+1}=\posenetpar{\wauxself;\ir{t},\ir{t+1}}$, between them. Then, as can be seen in \Fig{overall_training}, a projection module takes $\posetransfpar{t\rightarrow t+1}, \ir{t+1},$ and the depth estimation $\wmdepar{\ir{t}}$, to generate the synthesized frame $\irhpose{t+1 \rightarrow t}(\wmde,\wauxself)$ which, ideally, should match $\ir{t}$. In fact, both frames adjacent to $\ir{t}$ are considered for robustness. Thus, the SfM-inspired component of $\selfloss$ is defined as:
\begin{equation}
\minphotoerror{t}{\wmde,\wauxself} = \averagedcpe{\ir{t},\irhpose{t\pm1 \rightarrow t}(\wmde,\wauxself),\ir{t\pm1}} \enspace , \notag
\end{equation}
where $\cpepar{}$ is a pixelwise conditioned photo-metric error and $\averagedcpe{}$ its average over the pixels. Obtaining $\cpepar{}$ starts by computing two pixelwise photo-metric error measurements, $\pepar{\ir{t-1},\ir{t},\ir{t+1}}$ and $\pepar{\irhpose{t-1 \rightarrow t}(\wmde,\wauxself),\ir{t},\irhpose{t+1 \rightarrow t}(\wmde,\wauxself)}$, where $\pepar{\ir{-1},\ir{0},\ir{+1}} = \min(\pepar{\ir{0},\ir{-1}},\pepar{\ir{0},\ir{+1}})$, and $\pepar{\ir{A},\ir{B}}$ is the pixelwise photo-metric error between $\ir{A}$ and $\ir{B}$ defined in \cite{Godard:2017}, {\ie}, based on local structural similarity (SSIM) and pixelwise photo-metric absolute differences between $\ir{A}$ and $\ir{B}$. Thus, $\min()$ applies pixelwise. Then, a pixelwise binary \emph{auto-mask} \cite{Godard:2019MonoDepth2} is computed as:
\begin{align}
&\mr{t}{\ir{t},\irhpose{t\pm1 \rightarrow t}(\wmde,\wauxself),\ir{t\pm1}} = \notag \\
&\big[\pepar{\irhpose{t-1 \rightarrow t}(\wmde,\wauxself),\ir{t},\irhpose{t+1 \rightarrow t}(\wmde,\wauxself)} 
< \pepar{\ir{t-1},\ir{t},\ir{t+1}}\big]_{\mbox{{\scriptsize I}}} \notag ,
\end{align}
\noindent where $[\,\,]_{\mbox{{\scriptsize I}}}$ denotes the Iverson bracket applied pixelwise. Finally, $\cpepar{}$ is computed as:
\begin{align}
&\cpepar{\ir{t},\irhpose{t\pm1 \rightarrow t}(\wmde,\wauxself),\ir{t\pm1}} = \notag \\ 
&\mr{t}{\ir{t},\irhpose{t\pm1 \rightarrow t}(\wmde,\wauxself),\ir{t\pm1}} \odot  \notag \\ 
&\pepar{\irhpose{t-1 \rightarrow t}(\wmde,\wauxself),\ir{t},\irhpose{t+1 \rightarrow t}(\wmde,\wauxself)} \enspace , \notag 
\end{align}
\noindent where $\odot$ stands for pixelwise multiplication. The auto-mask $\mr{t}{}$ conditions which pixels of $\pepar{}$ are considered during the gradient computation of $\selfloss$. As explained in \cite{Godard:2019MonoDepth2}, the aim of $\mr{t}{}$ is to remove, during training, the influence of pixels which remain the same between adjacent frames because they are assumed to often indicate SfM violations such as a static camera, objects moving as the camera, or low texture regions. The support of $\wauxself$ is not needed at testing time.

\subsection{Supervised loss: $\suplosspar{\wmde; {\VWSequences.\VWSequencesGT}}$}
\label{ssec:suploss}
In this case, since we address an estimation problem and we have accurate GT, we base $\suploss$ on the L1 metric. On the other hand, MDE is specially interesting to determine how far is the ego-vehicle from vehicles, pedestrians, etc. Accordingly, since $\VWSequencesGT$ includes semantic class GT, we use it to increase the relevance of accurately estimating the depth for such major traffic protagonists. Moreover, since virtual-world depth maps are based on the Z-buffer involved on image rendering, the range of depth values available as GT tend to be over-optimistic even for active sensors such as LiDAR. For instance, there can be depth values larger than $300~m$ in the Z-buffer. Since we do not aim at estimating depth beyond a reasonable threshold (in $m$), $\maxestidepth$, to compute $\suploss$ we will also discard pixels $p$ with $\ds{t}(p)\geq\maxestidepth$. For each $\is{t}$, both the semantic class relevance and the out-of-range depth values, can be codified as real-valued weights running on $[0,1]$ and arranged on a mask, $\ms{t}$. Thus, $\ms{t}$ depends on $\ds{t}, \maxestidepth,$ and $\cs{t}$. However, contrarily to $\mr{t}{}$, we can compute $\ms{t}$ offline, {\ie}, before starting the training process. Taking all these details into account, we define our supervised loss as:
\begin{equation}
\label{eq:sup-loss}
\suplosspar{\wmde; {\VWSequences.\VWSequencesGT}} = \sum_{t=1}^{\Ns} \LONEMETRIC{\ms{t}\odot(\wmdepar{\is{t}}-\ds{t})} \enspace . 
\end{equation}

\subsection{Domain adaptation loss: $\dalosspar{\wmdeparenc, \wauxda; \RWSequences, \VWSequences}$}
\label{ssec:daloss}

As can be seen in \Fig{overall_training}, we aim at learning depth features, $\wmdeparenc$, so that it cannot be distinguished whether they were generated from a real-world input frame (target domain) or a virtual-world one (source domain); in other words, learning a domain invariant $\wmdeparenc$. Taking into account that we do not have accurate depth GT in the target domain, while we do have it for the source domain, we need to apply an unsupervised DA technique to train $\wmdeparenc$. In addition, as part of $\wmde$, the training of $\wmdeparenc$ must result on an accurate $\wmdepar{\inputimage}$. Achieving this accuracy and domain invariance are adversarial goals. Accordingly, we propose to use the Gradient-Reversal-Layer (GRL) idea introduced in \cite{Ganin:2015}, which, up to the best of our knowledge, has not been applied before for DA in the context of MDE. In this approach, the domain invariance of $\wmdeparenc$ is measured by a binary target/source domain-classifier CNN, $\danet$, of weights $\{\wmdeparenc,\wauxda\}$. In \cite{Ganin:2015}, a logistic loss is proposed to train the domain classifier. In our case, this is set as:
\begin{align}
\label{eq:da-loss}
&\dalosspar{\wmdeparenc, \wauxda;\RWSequences, \VWSequences} = \\ 
&\sum_{t=1}^{\Nr} \log(\danet(\wmdeparenc,\wauxda;\ir{t})) + \sum_{t=1}^{\Ns} \log(1-\danet(\wmdeparenc,\wauxda;\is{t})) 
\notag \enspace ,
\end{align}
\noindent where we assume that $\danet(\wmdeparenc,\wauxda;\inputimage)$ outputs 1 if $\inputimage\in\RWSequences$ and 0 if  $\inputimage\in\VWSequences$. The GRL has no parameters and connects $\wmdeparenc$ with $\wauxda$ (see \Fig{overall_training}). Its behavior is exactly as explained in \cite{Ganin:2015}. This means that during forward passes of training, it acts as an identity function, while, during back-propagation, it reverses the gradient vector passing through it. Both the GRL and $\wauxda$ are required at training time, but not at testing time.

\subsection{Overall training procedure} 
\label{ssec:traininprocedure}
Algorithm \ref{alg:gradientcomputation} summarizes the steps to compute the needed gradient vectors for mini-batch optimization. In particular, we need the gradients related to MonoDEVSNet weights, $\wmde=\{\wmdeparenc,\wmdeparpyr,\wmdepardec\}$, and the weights of the auxiliary tasks, {\ie}, $\wauxself$ for SfM self-supervision, and $\wauxda$ for DA. Regarding gradient computation, we do not need to distinguish $\wmdeparpyr$ from $\wmdepardec$, so we define $\wmdeparpyrdec=\{\wmdeparpyr,\wmdepardec\}$. In \Alg{gradientcomputation}, we introduce an equalizing factor between supervised and self-supervised losses, $\selflosseq\in\R$, which aims at avoiding one loss dominating over the other. A priori, we could set a constant factor. However, in practice, we have found that having an adaptive value is more useful. Therefore, inspired by the GradNorm idea \cite{Chen:2018GradNorm}, we use the ratio between the supervised and self-supervised losses. Algorithm \ref{alg:gradientcomputation} also introduces the scaling factor $\dalosseq\in\R$ which, following \cite{Ganin:2015}, controls the trade-off between optimizing $\wmdeparenc$ to obtain an accurate $\wmdepar{\inputimage}$ model versus being domain invariant. Finally, $\dalosspar{\wmdeparenc, \wauxda;\emptyset,\BVWSequences}$ and $\dalosspar{\wmdeparenc, \wauxda;\BRWSequences,\emptyset}$ indicate whether this loss must be computed only using virtual- or real-world data, respectively.  

%\setcounter{myalg}{0}
\begin{comment}
\begin{myalg}
\centering
\caption{Computing the gradients $\lossgradvalue{}{\wmdeparenc}$, $\lossgradvalue{}{\wmdeparpyrdec}$, $\lossgradvalue{}{\wauxself}$, $\lossgradvalue{}{\wauxda}$ for a mini-batch $\BRWSequences\subset\RWSequences, \BVWSequences.\BVWSequencesGT\subset\VWSequences.\VWSequencesGT$. $\lossgrad{\xi_i}F(\xi_1,\xi_2)$ refers to back-propagation on $F(\xi_1,\xi_2)$ with respect to weights $\xi_i\subset\xi_1\cup\xi_2$. $\emptyset$ is the empty set. $\selfloss$ is defined in \Eq{sf-loss}, $\suploss$ in \Eq{sup-loss}, and $\daloss$ in \Eq{da-loss}.}
\includegraphics[width=0.85\columnwidth]{gradientalgorithm.png}
\label{alg:gradientcomputation}
\end{myalg}
\end{comment}

{
\begin{algorithm}
\caption{Computing the gradients $\lossgradvalue{}{\wmdeparenc}$, $\lossgradvalue{}{\wmdeparpyrdec}$, $\lossgradvalue{}{\wauxself}$, $\lossgradvalue{}{\wauxda}$ for a mini-batch $\BRWSequences\subset\RWSequences, \BVWSequences.\BVWSequencesGT\subset\VWSequences.\VWSequencesGT$. $\lossgrad{\xi_i}F(\xi_1,\xi_2)$ refers to back-propagation on $F(\xi_1,\xi_2)$ with respect to weights $\xi_i\subset\xi_1\cup\xi_2$. $\emptyset$ is the empty set.}
\label{alg:gradientcomputation}
\end{algorithm}
\begin{minipage}{0.9\columnwidth}
{\centering Forward Passes with $\{\BVWSequences,\BVWSequencesGT\}$\par}
\vspace{-0.3cm}
\begin{align}
\ell^{\supervised}(\wmde)\gets&\suplosspar{\wmde;{\BVWSequences.\BVWSequencesGT}}\notag\\
\ell^{\domainadaptation,\syntheticdata}(\wmdeparenc,\wauxda)\gets&\dalosseq\dalosspar{\wmdeparenc,\wauxda;\emptyset,\BVWSequences}\notag
\end{align}
\\
{\centering Back-propagation for Supervision \& DA\par}
\vspace{-0.3cm}
\begin{align}
\lossgradvalue{\syntheticdata}{\wmdeparpyrdec}\gets&\lossgrad{\wmdeparpyrdec}\ell^{\supervised}(\wmde)\notag\\
\lossgradvalue{\syntheticdata}{\wmdeparenc}\gets&\lossgrad{\wmdeparenc}(\ell^{\supervised}(\wmde)\notag-\ell^{\domainadaptation,\syntheticdata}(\wmdeparenc,\wauxda))\notag\\
\lossgradvalue{\syntheticdata}{\wauxda}\gets&\lossgrad{\wauxda}\ell^{\domainadaptation,\syntheticdata}(\wmdeparenc,\wauxda)\notag
\end{align}
\\
{\centering Forward Passes with $\BRWSequences$\par}
\vspace{-0.3cm}
\begin{align}
\ell^{\selfsupervised}(\wmde,\wauxself)\gets&\selflosspar{\wmde,\wauxself;\BRWSequences}\notag\\
\ell^{\domainadaptation,\realdata}(\wmdeparenc,\wauxda)\gets&\dalosseq\dalosspar{\wmdeparenc,\wauxda;\BRWSequences,\emptyset}\notag
\end{align}
\\
{\centering Back-propagation for Self-supervision \& DA\par}
\vspace{-0.3cm}
\begin{align}
\lossgradvalue{\realdata}{\wmdeparpyrdec}\gets&\lossgrad{\wmdeparpyrdec}\ell^{\selfsupervised}(\wmde,\wauxself)\notag\\
\lossgradvalue{\realdata}{\wmdeparenc}\gets&\lossgrad{\wmdeparenc}(\ell^{\selfsupervised}(\wmde,\wauxself)-\ell^{\domainadaptation,\realdata}(\wmdeparenc,\wauxda))\notag\\
\lossgradvalue{\realdata}{\wauxda}\gets&\lossgrad{\wauxda}\ell^{\domainadaptation,\realdata}(\wmdeparenc,\wauxda)\notag
\end{align}
\\
{\centering Setting the final gradient vectors\par}
\vspace{-0.3cm}
\begin{align}
\lossgradvalue{}{\wauxself}\gets&\lossgrad{\wauxself}\ell^{\selfsupervised}(\wmde,\wauxself)\notag\\
\lossgradvalue{}{\wauxda}\gets&\lossgradvalue{\syntheticdata}{\wauxda} + \lossgradvalue{\realdata}{\wauxda}\notag\\
\selflosseq\gets&\ell^{\supervised}(\wmde)/\ell^{\selfsupervised}(\wmde,\wauxself)\notag\\
\lossgradvalue{}{\wmdeparpyrdec}\gets&\lossgradvalue{\syntheticdata}{\wmdeparpyrdec} + \selflosseq\lossgradvalue{\realdata}{\wmdeparpyrdec}\notag\\
\lossgradvalue{}{\wmdeparenc}\gets&\lossgradvalue{\syntheticdata}{\wmdeparenc} + \selflosseq\lossgradvalue{\realdata}{\wmdeparenc}\notag
\end{align}
\end{minipage}
%\end{table}
}

%%%%%%%%%%%%%%%%%%%%%%%%%%%%%%%%%%%%%%%%%%%%%%%%%%%%%%%%%%%%%%%%%%%%%
%% local settings
\sisetup{detect-weight,mode=text}
\renewrobustcmd{\bfseries}{\fontseries{b}\selectfont}
\renewrobustcmd{\boldmath}{}
% abbreviation
\newrobustcmd{\B}{\bfseries}
\newrobustcmd{\IL}{\underline}
% shorten the inter column spaces
\addtolength{\tabcolsep}{-4.1pt}
%%%%%%%%%%%%%%%%%%%%%%%%%%%%%%%%%%%%%%%%%%%%%%%%%%%%%%%%%%%%%%%%%%%%%

\subsection{Absolute depth computation} 
\label{ssec:scaling}

The virtual-world supervised data trains $\wmdepar{\inputimage}$ on absolute depth values, while the real-world SfM self-supervised data trains $\wmdepar{\inputimage}$ on relative depth values. Thanks to the unsupervised DA, the depth features $\wmdeparenc$ are trained to be domain invariant. However, according to our experiments, this is not sufficient for $\wmdepar{\inputimage}$ producing accurate absolute depth values at testing time. Fortunately, thanks to the use of virtual-world data, we can still compute a global scaling factor, $\depthscalefactor\in\R$, so that $\depthscalefactor\wmdepar{\inputimage}$ is accurate in absolute depth terms. For that, we assume that the sequences in $\VWSequences$ are acquired with a camera analogous to the one used to acquire the sequences in $\RWSequences$. Here analogous refers to using the same number of pixels, field of view, frame rate, and mounted on-board in similar heading directions. Note that simulators are flexible enough for setting these camera parameters as needed. Accordingly, we train a $\wmdepar{\inputimage}$ model using only data from $\VWSequences$ and SfM self-supervision, {\ie} as if we would not have supervision for $\VWSequences$. Then, we find the median depth value produced by this model on the virtual-world data, $\meanestidepth\in\R$. Finally, we set $\depthscalefactor=\meangtdepth/\meanestidepth$, where $\meangtdepth\in\R$ is the median depth value of the GT. This pre-processing step is performed once and the model discarded afterwards. Other works apply a similar approach \cite{Zhou:2017, Godard:2019MonoDepth2, Cheng:2020S3Net, Guizilini:2020semantic, Zhao:2020} but relying on LiDAR data as GT reference, while we only rely on virtual-world data.

\section{Experimental Results}
\label{sec:experiments}
We start by defining the datasets and evaluation metrics used in our experiments. After, we provide relevant implementation and training details of MonoDEVSNet. Finally, we present and discuss our quantitative and qualitative results, comparing them with those from previous literature as well as performing an ablative analysis over MonoDEVSNet components.

\subsection{Datasets and evaluation metrics}
\label{ssec:datasets}

We use publicly available datasets and metrics which are \emph{de facto} standards in MDE research. In particular, we use KITTI Raw (KR) \cite{Geiger:2013} and Virtual KITTI (VK) \cite{Cabon:2020} as real- and virtual-world sequences, respectively. We follow Zhou {\etal} \cite{Zhou:2017} training-testing split. From the training split we select 12K monocular triplets, {\ie}, samples of the form $\{\ir{t-1}, \ir{t}, \ir{t+1}\}$. The testing split consists of 697 isolated images with LiDAR-based GT, actually introduced by Eigen {\etal} \cite{Eigen:2014}. In addition, for considering the semantic content of the images in the analysis of results, we also use KITTI Stereo 2015 (KS) \cite{Menze:2015} for testing. This dataset consists of 200 isolated images with enhanced depth maps and semantic labels. VK is used only for training, we also use 12K monocular triplets (non-rotated camera subset) with associated depth GT. In this case, the triplets are used to calibrate the global scaling factor $\depthscalefactor$ (see \sSect{scaling}), while for actual training supervision only 12K isolated frames are used. As the depth GT of VK ranges up to $\sim655$m, to match the range of KR's LiDAR-based GT, we clip it to $80$m ($\maxestidepth$). VK includes similar weather conditions as KR/KS, and adds situations with fog, overcast, and rain, as well as sunrise and sunset illumination. 

As is common practice since \cite{Godard:2017}, we use Make3D dataset \cite{Saxena:2009} for assessing generalization. It contains photographs of urban and natural areas. Thus, Make3D shows views and content pretty much different from those on-board a vehicle as KR, KS, and VK. The images come with depth GT acquired by a 3D scanner. There are 534 images with depth GT, organized in a standard split of 400 for training and 134 for testing. We use the latter, since we rely on Make3D only for testing. 

In order to assess quantitative MDE results, we use the standard metrics introduced by Eigen {\etal} \cite{Eigen:2014}, {\ie}, the average absolute relative error (abs-rel), the average squared relative error (sq-rel), the root mean square error (rms), and the rms log error (rms-log). For these metrics, the lower the better. In addition, the accuracy (termed as $\delta$) under a threshold $\tau\in\{1.25, 1.25^2, 1.25^3\}$ is also used as metric. In this case, the higher the better. The abs-rel error and the ${\delta<\tau}$ are percentage measurements, sq-rel and rms are reported in meters, and rms-log is similar (reported in meters) to rms but applied to logarithm depth values. 

These metrics are applied to absolute depth values for MDE models trained with depth supervision coming from either LiDAR  \cite{Eigen:2014, Liu:2016, Roy:2016, Laina:2016, Cao:2017, Fu:2018DORN, Gurram:2018, He:2018, Xu:2018, Yin:2019, Guizilini:2020}, stereo \cite{Saxena:2007, Garg:2016, Godard:2017, Godard:2019MonoDepth2, Pillai:2019}, real-world stereo and virtual-world depth \cite{Zhao:2019GASDA, Pnvr:2020SharinGAN}, or stereo and LiDAR \cite{Kuznietsov:2017, He:2018wearable}. 
However, MDE models trained on pure SfM self-supervision can only estimate depth in relative terms, {\ie}, up to scale. Moreover, the scale factor varies from image to image, a problem known as scale inconsistency. In this case, before computing the above metrics, it is applied a per-image correction factor computed at testing time \cite{Zhou:2017, Yin:2018GeoNet, Zhao:2020, Godard:2019MonoDepth2, Guizilini:2020semantic, Cheng:2020S3Net}. In particular, given a test image $\inputimage$ with GT and estimated depth $d(\inputimage)$ and $\hat{d}(\inputimage)$, respectively, the common practice consists of computing a scale $\depthscalefactor(\inputimage) \in \R$ as the ratio $\mbox{median}(d(\inputimage))/\mbox{median}(\hat{d}(\inputimage))$, and then compare $\depthscalefactor(\inputimage)\hat{d}(\inputimage)$ with $d(\inputimage)$. On the other hand, SfM self-supervision with the help of additional information can train models able to produce absolute scale in testing time. For instance, \cite{Guizilini:20203D} uses the ego-vehicle speed and, in fact, virtual-world supervision can help too \cite{Zheng:2018T2Net, Kundu:2018AdaDepth}. The latter approach is the one followed in this paper, especially thanks to the procedure presented in \sSect{scaling}. Therefore, $\wmdepar{\inputimage}$ will be evaluated in relative scale terms, and $\depthscalefactor\wmdepar{\inputimage}$ in absolute terms. Please, note that our $\depthscalefactor\in\R$ scaling factor is constant for all the evaluated images and computed at training time. In the following, when presenting quantitative results, we will make clear if they are in relative or absolute terms.

\subsection{Implementation details}
\label{ssec:implementation}

\begin{figure}
    \centering
    \includegraphics[width=0.8\columnwidth]{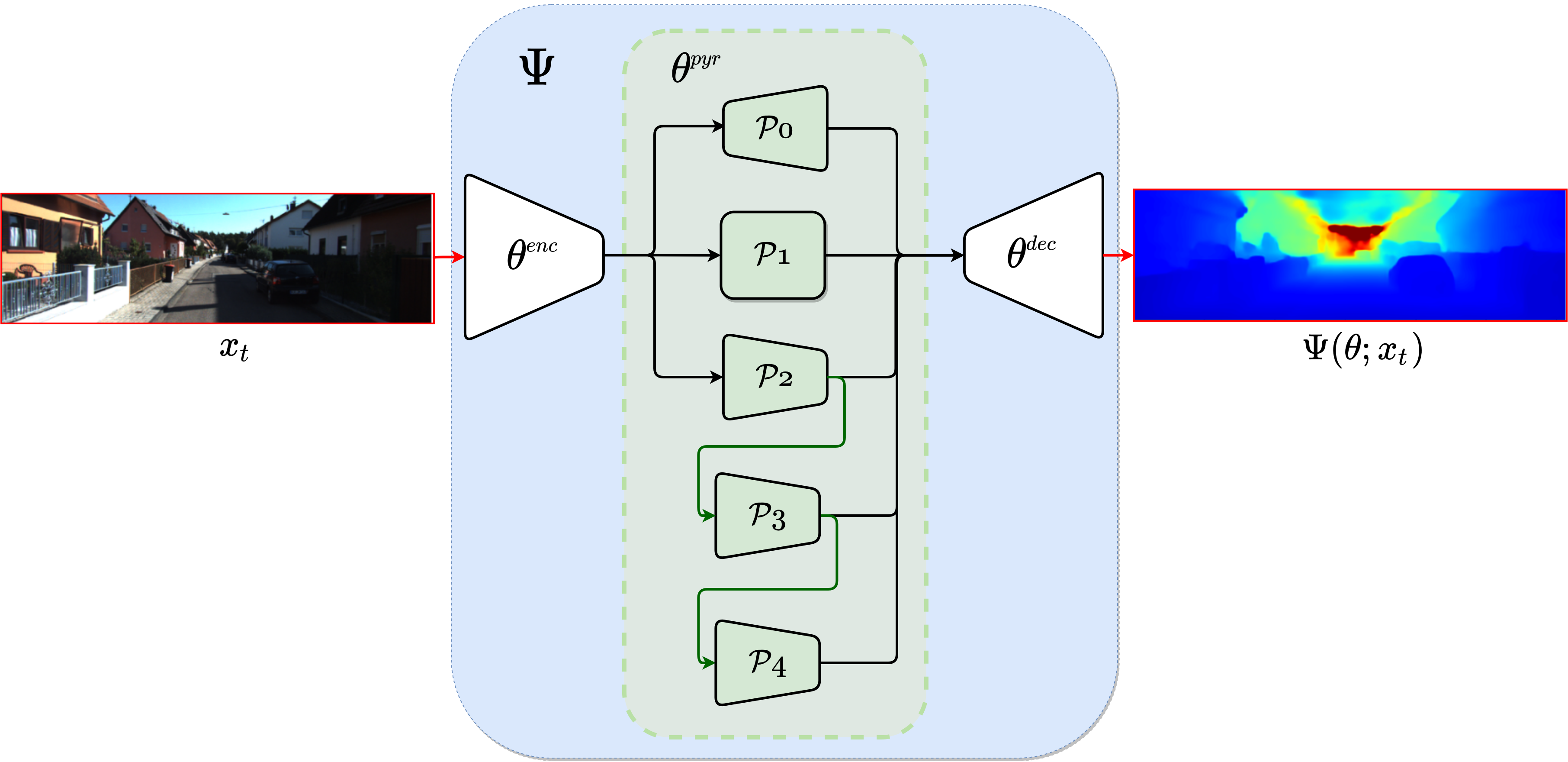}
    \caption{Pyramidal architecture of $\wmdeparpyr$.}
    \label{fig:mde_net_arch}
\end{figure}

We start by selecting the actual CNN layers to implement $\wmdepar{\inputimage}$. Since we leverage the SfM self-supervision idea from \cite{Godard:2019MonoDepth2}, a straightforward implementation would be to use its ResNet-based architecture as it is. However, the High-Resolution Network (HRNet) architecture \cite{Wang:2020HrNet}, exhibits better accuracy in visual tasks such as semantic segmentation and object detection, suggesting that it can be a better backbone than ResNet. Thus, we decided to start our experiments by comparing ResNet and HRNet backbones using the SfM self-supervision framework provided in \cite{Godard:2019MonoDepth2}. In particular, we assess different ResNet/HRNet architectures for $\wmdeparenc$, while using the proposal in \cite{Godard:2019MonoDepth2} for $\wmdepardec$. Then, when using ResNet we have $\wmdeparpyr=\emptyset$, while for HRNet $\wmdeparpyr$ consists of pyramidal layers adapting the $\wmdeparenc$ and $\wmdepardec$ CNN architectures under test. For these experiments, we rely on KR. Table \ref{tab:compare_netarch} shows the accuracy (in relative scale terms) of the tested variants and their number of weights. We see how HRNet outperforms ResNet, being HRNet-W48 the best. Indeed, HRNet is slower than ResNet, and HRNet-W48 is the one requiring more GFLOPS by far. However, at this stage of our research we target the architecture which potentially can provide higher depth estimation accuracy. Thus, for our following experiments, we will rely on HRNet-W48 although being the heaviest. We show the corresponding pyramidal architecture of $\wmdeparpyr$ in \Fig{mde_net_arch}. It is composed of five blocks ($\mathcal{P}_i$), where each block is a pipeline of three consecutive layers consisting of convolution, batch normalization, and ReLU. As a deep learning framework, we use PyTorch 1.5v \cite{Paszke:2019pytorch}.

\begin{table}
\centering
\caption{Comparing ResNet and HRNet as backbone for $\wmdepar{\inputimage}$, training only on SfM self-supervision (relative scale) using the framework in \cite{Godard:2019MonoDepth2}. {MW} column stands for millions of $\wmdeparenc$ weights to be learnt. 
{FPS} stands for \emph{frames per second} as required by $\wmdepar{\inputimage}$ to process $\inputimage$, while {GFLOPS} refers to the \emph{giga floating-point operations per second} required by $\wmdeparenc$; in both cases using an NVIDIA RTX 2080Ti GPU. The $1.25^n$ columns, $n\in\{1,2\}$, refer to the $\tau$ in the usual $\delta<\tau$ accuracy metrics. In all the tables of \Sect{experiments}, bold stands for {\bf best} and underline for \IL{second-best}. $(^*)$ Currently, HRNet branches do not run in parallel in PyTorch, thus, compromising speed.}
\label{tab:compare_netarch}
\begin{tabular}{|c||*{10}{c|}}\hline
\makebox[5em]{$\wmdeparenc$ Backb.}
&\makebox[2em]{MW}&\makebox[4em]{GFLOPS}&\makebox[2.5em]{FPS} &\makebox[2.5em]{abs-rel}&\makebox[2.5em]{sq-rel}&\makebox[2em]{rms}&\makebox[2.5em]{$1.25$}&\makebox[2.5em]{$1.25^2$}\\\hline \hline
ResNet-18       & \IL{11.6} & \B 4.47   & \B 141.2   &  0.115     & 0.882      & 4.701      & 0.879      & 0.961 \\
ResNet-50       & 25.5      & 10.14     & \IL{77.06} & 0.110      & 0.831      & 4.642      & 0.883      & 0.962 \\
ResNet-101      & 44.5      & 19.29     &     43.26  & 0.110      & 0.809      & 4.712      & 0.878      & 0.960 \\
ResNet-152      & 60.2      & 28.47     &     30.71  & 0.107      & \IL{0.800} & \IL{4.629} & 0.885      & 0.962  \\ \hline
HRNet-W18       & \B 9.5    & \IL{8.29} &     $15.79^*$  & \IL{0.107} & 0.846      & 4.671      & \IL{0.887} & \IL{0.962} \\ 
HRNet-W32       & 29.3      & 19.50     &     $15.53^*$  & 0.107      & 0.881      & 4.794      & 0.886      & 0.961 \\
HRNet-W48       & 65.3      & 40.04     &     $15.48^*$  & \B 0.105   & \B 0.791   & \B 4.590   & \B 0.888   & \B 0.963 \\\hline
\end{tabular}
\end{table}

In order to train the camera pose estimation network, $\posenetpar{\wauxself;\ir{t},\ir{t\pm1}}$, we follow \cite{Godard:2019MonoDepth2} but
using ResNet-50 instead of ResNet-18 since the former is more accurate. Four convolutional layers are used to convert the ResNet-50 bottleneck features to the 6-DoF relative pose vector (3D translation and rotation). For training the classification block of $\danet(\wmdeparenc,\wauxda;\inputimage)$, {\ie}, $\wauxda$, we use a standard classification pipeline based on convolutions, ReLU and fully connected layers. Finally, we remark that these networks are not required at testing time.

\subsection{Training details}
\label{ssec:training}

The input images are processed (at training and testing time) at a resolution of ${640\times192}$ ({$W \times H$}), where LANCZOS interpolation is performed from the ${\sim1242\times375}$ original resolution. As optimizer, we use Adam \cite{Kingma:2015} with learning rate ${lr=10^{-4}}$, and the rest of its hyper-parameters set to default values. The weights $\wmdeparenc$ are initialized from available ImageNet \cite{Deng:2009} pre-training, $\wmdeparpyr, \wmdepardec$, and $\wauxda$ are randomly initialized with Kaiming weights, while the ResNet-50 part of $\wauxself$ is also initialized with ImageNet and the rest (convolutional layers to output the pose vector) following Kaiming. The mini-batch size is of 16 images, 50\%/50\% from real/virtual domains. To minimize over-fitting, we apply standard data augmentation such as horizontal flip, a 50\% chance of random brightness, contrast, saturation, and hue jitter with ranges of $\pm0.2$, $\pm0.2$, $\pm0.2$, and $\pm0.1$, respectively. Remaining hyper-parameters were set as $\lambda=0.001$ in \Eq{sf-loss}, $\dalosseq=10$ in \Alg{gradientcomputation}, and in \Eq{sup-loss} our mask $\ms{t}$ is set to have values of $1.0$ for traffic participants (vehicles, pedestrians, {\etc}), $0.5$ for static infrastructure (buildings, road, vegetation, {\etc}), and $0.0$ for the sky and pixels with depth over $\maxestidepth$ (here $80$m).

\subsection{Results and discussion}
\label{ssec:results}

\subsubsection{Relative depth assessment}
We start by assessing MDE in relative terms. Table \ref{tab:SOTA_KITTI_eigen_relative} presents MonoDEVSNet results (Ours) and those from previous works based on SfM self-supervision. From this table we can draw several observations. Regarding DA, MonoDEVSNet (VK\_v1) outperforms $S^3$Net (VK\_v1) in all metrics. The new version of VK (VK\_v2) allows us to obtain even better results. MonoDEVSNet with virtual-world supervision outperforms the version with only SfM self-supervision (best result in \Tab{compare_netarch}) in all metrics, no matter the VK version we use. Overall, MonoDEVSNet outperforms most previous methods, being on pair with  \cite{Guizilini:2020semantic}.

\begin{table}
\centering
\caption{\emph{\underline{Relative depth}} results up to $80$m on the (KR) Eigen {\etal} \cite{Eigen:2014} testing split. These methods rely on SfM self-supervision. In addition, methods in gray use DA supported by VK. ${(^1)}$ MonoDepth2 is based only on SfM self-supervision.}
\label{tab:SOTA_KITTI_eigen_relative} 
\begin{tabular}{|l||*{7}{c|}}\hline
Method &\makebox[2em]{abs-rel}&\makebox[2em]{sq-rel}&\makebox[2em]{rms}&\makebox[3em]{rms-log}&\makebox[2.25em]{$1.25$}&\makebox[2.25em]{$1.25^2$}&\makebox[2.25em]{$ 1.25^3$}\\\hline \hline
\cite{Zhou:2017} (Zhou {\etal})            & 0.183      & 1.595      & 6.709      & 0.270      & 0.734      & 0.902      & 0.959 \\ \hline
\cite{Yin:2018GeoNet} GeoNet               & 0.149      & 1.060      & 5.567      & 0.226      & 0.796      & 0.935      & 0.975 \\ \hline
\cite{Godard:2019MonoDepth2}  MonoDepth2$^1$   
                                           & 0.115      & 0.903      & 4.863      & 0.193      & 0.877      & 0.959      & 0.981 \\ \hline
\cite{Zhao:2020} (Zhao {\etal})            & 0.113      & 0.704      & 4.581      & 0.184      & 0.871      & 0.961      & \B 0.984 \\ \hline
\cite{Guizilini:2020semantic} (Guizilini {\etal}) 
                                           & \B 0.102    & \IL{0.698} & \IL{4.381} & \B 0.178   & \B 0.896   & 0.964      & \B 0.984 \\ \hline
\rowcolor{Gray}
\cite{Cheng:2020S3Net} $S^3$Net (VK\_v1)   & 0.124       & 0.826      & 4.981      & 0.200      & 0.846      & 0.955      & 0.982   \\ \hline
\rowcolor{Gray}
MonoDEVSNet / VK\_v1                              & \IL{0.105}  &  0.753     & 4.389      & \IL{0.179} & 0.890      & \IL{0.965} & \IL{0.983} \\ \hline
\rowcolor{Gray}
MonoDEVSNet / VK\_v2                              & \B 0.102    & \B 0.685   & \B 4.303   & \B 0.178   & \IL{0.894} & \B 0.966   & \B 0.984 \\ \hline
\end{tabular}
\end{table}

\subsubsection{Absolute depth assessment}
While assessing depth in relative terms is a reasonable option to compare methods purely based on SfM self-supervision, the most relevant evaluation is in terms of absolute depth. These are presented in \Tab{SOTA_KITTI_eigen_absolute}. The first (top) block of this table shows results based on depth supervision from LiDAR, thus, a priori they can be thought of as upper-bounds for methods based on self-supervision. The second block shows methods that only use virtual-world supervision. The third and fourth (bottom) blocks show results based on stereo and SfM self-supervision, respectively. Methods in gray use DA supported by VK. We can draw several observations from this table. MonoDEVSNet (Ours) is the best performing among those leveraging supervision from VK\_v1 and, consistently with the results on relative depth, by using VK\_v2 we improve MonoDEVSNet results. In fact, MonoDEVSNet based on VK\_v2 outperforms all self-supervised methods, including those using stereo rigs instead of monocular systems. We are not yet able to reach the performance of the best methods supervised with LiDAR data. However, it is clear that our proposal is able to successfully combine real-world SfM self-supervision and virtual-world supervision. Thus, we think it is worth to keep this line of research until reaching the LiDAR-based upper-bounds.

\begin{table}[!t]
\centering
\caption{\emph{\underline{Absolute depth}} results up to $80$m on the (KR) Eigen {\etal} \cite{Eigen:2014} testing split. We divide the results into four blocks. From top to bottom, the blocks refer to: methods based on LiDAR supervision, only virtual-world supervision, stereo self-supervision, SfM self-supervision. In these blocks, we remark best and second-best results per block. Methods in gray use DA supported by VK. We remark some additional comments: ${(^1)}$ in addition to LiDAR supervision, it also uses stereo self-supervision; ${(^2)}$ it uses stereo and SfM self-supervision; ${(^3)}$ in this case, the MDE network is pre-trained on Cityscapes dataset \cite{Cordts:2016} and then fine-tuned on KITTI.}
\label{tab:SOTA_KITTI_eigen_absolute} 
\begin{tabular}{|l||*{7}{c|}}\hline
Method &\makebox[2em]{abs-rel}&\makebox[1.9em]{sq-rel}&\makebox[1.9em]{rms}&\makebox[2.9em]{rms-log}&\makebox[2.1em]{$1.25$}&\makebox[2.1em]{$1.25^2$}&\makebox[2.1em]{$ 1.25^3$}\\\hline \hline
\cite{Eigen:2014} (Eigen {\etal})           & 0.203      & 1.548      & 6.307      & 0.282      & 0.702      & 0.890      & 0.890 \\ \hline
\cite{Liu:2016} (Liu {\etal})	            & 0.217	     & 1.841      & 6.986	   & 0.289	    & 0.647	     & 0.882      & 0.961 \\ \hline
\cite{Cao:2017} (Cao {\etal})               & 0.115	     & N/A        & 4.712      & 0.198	    & 0.887      & 0.963      & 0.982 \\ \hline
\cite{Kuznietsov:2017} (Kuzni. {\etal})$^1$ & 0.113	     & 0.741	  & 4.621	   & 0.189	    & 0.862	     & 0.960	  & 0.986 \\ \hline
\cite{Xu:2018} (Xu {\etal})                 & 0.122      & 0.897      & 4.677      & N/A        & 0.818      & 0.954      & 0.985 \\ \hline
\cite{Gurram:2018} (Gurram {\etal})         & 0.100      & \IL{0.601} & 4.298      & 0.174      & 0.874      & 0.966      & \IL{0.989} \\ \hline
\cite{Fu:2018DORN} DORN                     & \IL{0.098} & \B 0.582   & \IL{3.666} & \IL{0.160} & \IL{0.899} & \IL{0.967} & 0.986  \\ \hline
\cite{Yin:2019} VNL                       & \B 0.072   & N/A        & \B 3.258   & \B 0.117   & \B 0.938   & \B 0.990   & \B 0.998 \\ \hline
\Xhline{4\arrayrulewidth}
\rowcolor{Gray}
\cite{Kundu:2018AdaDepth} AdaDepth / VK\_v1 & \B 0.167   & \B 1.257   & \B 5.578   & \B 0.237   & \B 0.771   & \B 0.922   & \B 0.971 \\ \hline
\rowcolor{Gray}
\cite{Zheng:2018T2Net} $T^2$Net / VK\_v1       & \IL{0.174} & \IL{1.410} & \IL{6.046} & \IL{0.253} & \IL{0.754} & \IL{0.916} & \IL{0.966} \\ \hline
\Xhline{4\arrayrulewidth}
\cite{Garg:2016} (Garg {\etal}) 	        & 0.169	     & 1.512	  & 5.763	   & 0.236	    & 0.836	     & 0.935	  & 0.968 \\ \hline
\cite{Pillai:2019} SuperDepth               & 0.112      & 0.875      & \IL{4.958} & \IL{0.207} & 0.852      & 0.947      & 0.977 \\ \hline
\cite{Godard:2019MonoDepth2} MonoDepth2     & \IL{0.109} & \IL{0.873} & 4.960      & 0.209      &\IL{0.864}  & \IL{0.948} & 0.975 \\ \hline
\cite{Godard:2019MonoDepth2} MonoDepth2$^2$ & \B 0.106   & \B 0.806   & \B 4.630   & \B 0.193   & \B 0.876   & \B 0.958   & \B 0.980 \\ \hline
\rowcolor{Gray}
\cite{Zhao:2019GASDA} GASDA / VK\_v1        & 0.120      & 1.022      & 5.162      & 0.215      & 0.848      & 0.944      & 0.974 \\ \hline
\rowcolor{Gray}
\cite{Pnvr:2020SharinGAN} SharinGAN / VK\_v1& 0.116      & 0.939      & 5.068      & 0.203      & 0.850      & \IL{0.948} & \IL{0.978} \\ \hline
\Xhline{4\arrayrulewidth}
\cite{Guizilini:20203D} PackNet-SfM         & 0.111      & 0.829      & 4.788      & 0.199      & 0.864      & 0.954      & 0.980 \\ \hline
\cite{Guizilini:20203D} PackNet-SfM$^3$     & \IL{0.108} & 0.803      & 4.642      & 0.195      & \IL{0.875} & 0.958      & 0.980 \\ \hline
\rowcolor{Gray}
MonoDEVSNet / VK\_v1                               & \IL{0.108} & \IL{0.775} & \IL{4.464} & \IL{0.188} & \IL{0.875} & \IL{0.961} & \IL{0.982} \\ \hline 
\rowcolor{Gray}
MonoDEVSNet / VK\_v2                               & \B 0.104   & \B 0.721   & \B 4.396   & \B 0.185   & \B 0.880   & \B 0.962   & \B 0.983 \\ \hline
\Xhline{4\arrayrulewidth}
\end{tabular}
\end{table}
\begin{table}[!h]
\centering
\caption{\emph{\underline{Absolute depth}} ablative results of MonoDEVSNet (VK\_v2) up to $80$m on the (KR) Eigen testing split \cite{Eigen:2014}. Rows 1-6 show the progressive use of the components of our proposal (each row adds a new component). ${50/50}$ refers to mini-batches of 50\% real-world samples and 50\% or virtual-world ones; not using ${50/50}$ (rows 1-2) means that we alternate mini-batches of pure real- or virtual-world samples. Row 7 corresponds to a simplification of the SfM self-supervised loss. $\wauxgan$ (rows 8-9) refers to a GAN-based DA approach. LB (lower bound, row 10) indicates the use of only virtual-world data. UB (upper bound, row 12) indicates the use of KITTI LiDAR-based supervision instead of virtual-world data. Rows 11 and 13 show the difference of our best model (All) with respect to LB and UB, respectively. ${\uparrow D}$ means that All is $D$ units better, while ${\downarrow D}$ means that it is $D$ units worse. All/W18 (row 14) and All/W32 (row 15) refer to using the All configuration by relying on HRNet-W18 and HRNet-W32, respectively.
} 
\label{tab:Ablative_KITTI_eigen_absolute} 
\begin{tabular}{|l||*{7}{c|}}\hline
Configuration &\makebox[2em]{abs-rel}&\makebox[2em]{sq-rel}&\makebox[2em]{rms}&\makebox[2.8em]{rms-log}&\makebox[2em]{$1.25$}&\makebox[2em]{$1.25^2$}&\makebox[2em]{$ 1.25^3$}\\\hline \hline
1.~$\{\RWSequences,\VWSequences.\VWSequencesGT\}$           & 0.368    & 2.601    & 8.025    & 0.514    & 0.080    & 0.478    & 0.883 \\ \hline
2.~$+\depthscalefactor$                                     & 0.140    & 0.876    & 4.915    & 0.217    & 0.828    & 0.950    & 0.980 \\ \hline
3.~$+{50/50}$                                               & 0.128    & 0.880    & 4.618    & 0.198    & 0.844    & 0.957    & 0.982 \\ \hline
4.~$+\wauxda$                                               & 0.110    & 0.724    & 4.450    & 0.187    & 0.873    & 0.960    & 0.983 \\ \hline
5.~$+\selflosseq$                                           & 0.106    & \B 0.716    & 4.441    & 0.188    & 0.876    & 0.962    & 0.982 \\ \hline
%5.1$\depthscalefactor, {50/50}, \wauxda, \ms{t} $           & 0.116    & 0.748    & 4.442    & 0.194    & 0.871    & 0.960    & 0.982 \\ \hline
6.~$+\ms{t}$ (\textbf{All})                                 & \B 0.104 & 0.721 & \B 4.396 & \B 0.185 & \B 0.880 & \B 0.962 & \B 0.983 \\ \hline
7.~Simplified $\selfloss$                                  & 0.105    & 0.736    &	4.471    & 0.190    & 0.875    & 0.960    &	0.981   \\ \hline
\Xhline{4\arrayrulewidth}
8.~All$+\wauxgan ; -\wauxda$                                & 0.119    & 0.809    & 4.654    & 0.196    & 0.857    & 0.958    & 0.982 \\ \hline
9.~All$+\wauxgan$                                           & 0.106    & 0.748    & 4.503    & 0.191    & 0.873    & 0.959    & 0.981 \\ \hline
\Xhline{4\arrayrulewidth}
10.~LB                                                       & 0.165    & 1.280    & 5.628    & 0.248    & 0.777    & 0.916    & 0.965   \\ \hline
11.~{{$\uparrow$All} {\vs} {$\downarrow$LB}}                & {$\uparrow$\itn{0.061}}   & {$\uparrow$\itn{0.559}}   & {$\uparrow$\itn{1.232}}   & {$\uparrow$\itn{0.063}}   & {$\uparrow$\itn{0.103}}    & {$\uparrow$\itn{0.046}}    & {$\uparrow$\itn{0.018}}   \\ \hline
12.~UB                                                      & 0.088    & 0.583    & 3.978    & 0.164    & 0.906    & 0.970    & 0.986   \\ \hline
13.~{{$\uparrow$All} {\vs} {$\downarrow$UB}}                & {$\downarrow$\itn{0.016}}   & {$\downarrow$\itn{0.138}}   & {$\downarrow$\itn{0.418}}   & {$\downarrow$\itn{0.021}}   & {$\downarrow$\itn{0.026}}    & {$\downarrow$\itn{0.008}}    & {$\downarrow$\itn{0.003}}   \\ \hline
\Xhline{4\arrayrulewidth}
14.~All/W18                                                 & 0.109     & 0.773     & 4.524     & 0.190 & 0.871 & 0.960 & 0.982 \\ \hline
15.~All/W32                                                 & 0.107     & 0.754     & 4.510     & 0.188 & 0.875 & 0.960 & 0.982 \\ \hline \Xhline{4\arrayrulewidth}
\end{tabular}
\end{table}

\subsubsection{Ablative analysis of MonoDEVSNet}
It is also worth to analyze the contribution of the main components of our proposal. In rows 1-6 of \Tab{Ablative_KITTI_eigen_absolute}, we add one component at a time showing performance for absolute depth. The 1st row corresponds to using the real-world data with SfM self-supervision and the virtual-world images with only depth supervision, {\ie}, without using neither semantic supervision ($\ms{t}$), nor gradient equalization ($\selflosseq$), nor domain adaptation ($\wauxda$), nor mixed mini-batches (${50/50}$), nor the global scaling factor ($\depthscalefactor$). By comparing 1st and 2nd rows ({\ie}, w/o $\depthscalefactor$ and w/ $\depthscalefactor$, resp.), we can see how relevant is obtaining a good global scaling factor to output absolute depth. In fact, adding $\depthscalefactor$ to the virtual-world depth supervision shows the higher improvement among all the components of our proposal. Then, using mixed mini-batches of real- and virtual-world data improves the performance over alternating mini-batches of only either real- or virtual-world data. This can be seen by comparing 2nd and 3rd rows ({\ie}, w/o ${50/50}$ and w/ ${50/50}$, resp.). If we alternate the domains, the optimization of a mini-batch is dominated by self-supervision (real-world data), and the optimization of the next mini-batch is dominated by supervision (virtual-world data). Thus, there is not an actual joint optimization of SfM self-supervised and supervised losses, which turns to be relevant. Yet, as can be seen in 4th row, when we add the DA component ($\wauxda$) we improve further the depth estimation results. As can bee seen in 5th row, adding the equalization ($\selflosseq$) between gradients coming from supervision and self-supervision also improves the depth estimation results. Finally, adding the virtual-world mask ($\ms{t}$) leads to the best performance in 6th row. Overall, this analysis shows how all the considered components are relevant in our proposal. We also remark that these components are needed only to train $\wmde$, but only $\depthscalefactor$ and $\wmde$ are required at testing time. Additionally, we have assessed the effect of simplifying the SfM self-supervised loss that we leverage from \cite{Godard:2019MonoDepth2}, here summarized in \sSect{slfloss}. In particular, we neither use the auto-mask ($\mr{t}{}$), nor the multi-scale depth loss, and we replaced the minimum re-projection loss by the usual average re-projection loss ({\ie}, we re-define $\pepar{\ir{-1},\ir{0},\ir{+1}}$ in \sSect{slfloss}). Results are shown in the 7th row. The metrics show worse values than in 6th row (All), but still outperforming or being on pair with PackNet-SfM and the stereo self-supervised methods of \Tab{SOTA_KITTI_eigen_absolute}. 

We also did additional experiments changing the DA mechanism. Instead of taking direct real- and virtual-world images as input to train $\wmdepar{\inputimage}$, a GAN-based CNN, $\gan$, processes them to create an image space in which (hopefully) it is not possible to distinguish the domain. We train a CNN, $\wmdepar{\ganpar{\wauxgan;\inputimage}}$, where $\inputimage$ can come from either the real or the virtual domain, and $\wauxgan$ are the weights of $\gan$. These weights are jointly trained with $\wmde, \wauxself,$ and $\wauxda$ to optimize depth estimation and minimize the possibility of discriminating the original domain of a sample $\igan=\ganpar{\wauxgan;\inputimage}$. Table \ref{tab:Ablative_KITTI_eigen_absolute} shows results using this GAN when removing $\wauxda$ (8th row) and when keeping it (9th row). As we can see, this approach does not improve performance. Moreover, the training is more complex and $\ganpar{\wauxgan;\inputimage}$ would be required at testing time. Thus, we discarded it.

We also assessed the improvement of our proposal with respect a lower-bound model (LB) trained on virtual-world images and their depth GT ($\VWSequences.\VWSequencesGT$), but neither using real-world data ($\RWSequences$), nor DA ($\wauxself$), nor the mask ($\ms{t}$). Results are shown in 10th row of \Tab{Ablative_KITTI_eigen_absolute}, and we explicitly show the improvement of our proposal over such LB in 11th row. Likewise, we have trained an upper-bound model (UB) replacing VK data by KR data with LiDAR-based supervision, so that DA is not required. Results are shown in 12th row, and the distance of our model to this UB is explicitly shown in 13th row. Comparing 11th and 13th rows we can see how we are clearly closer to the UB than to the LB. 

Finally, we have done experiments using HRNet-W18 and HRNet-W32. The results are shown in 14th and 15th rows of \Tab{Ablative_KITTI_eigen_absolute}, respectively. Indeed, as it happens with the results on relative depth (\Tab{compare_netarch}), HRNet-W48 outperforms these more lightweight versions of HRNet. However, by using HRNet-W18 and HRNet-W32 we still outperform or are on pair with the state-of-the-art self-supervised methods shown in \Tab{SOTA_KITTI_eigen_absolute}, {\ie}, those based on stereo self-supervision and PackNet-SfM.

\begin{figure*}
    \centering
    \includegraphics[% clip=true,trim=0 197 0 0,
    width=\textwidth]{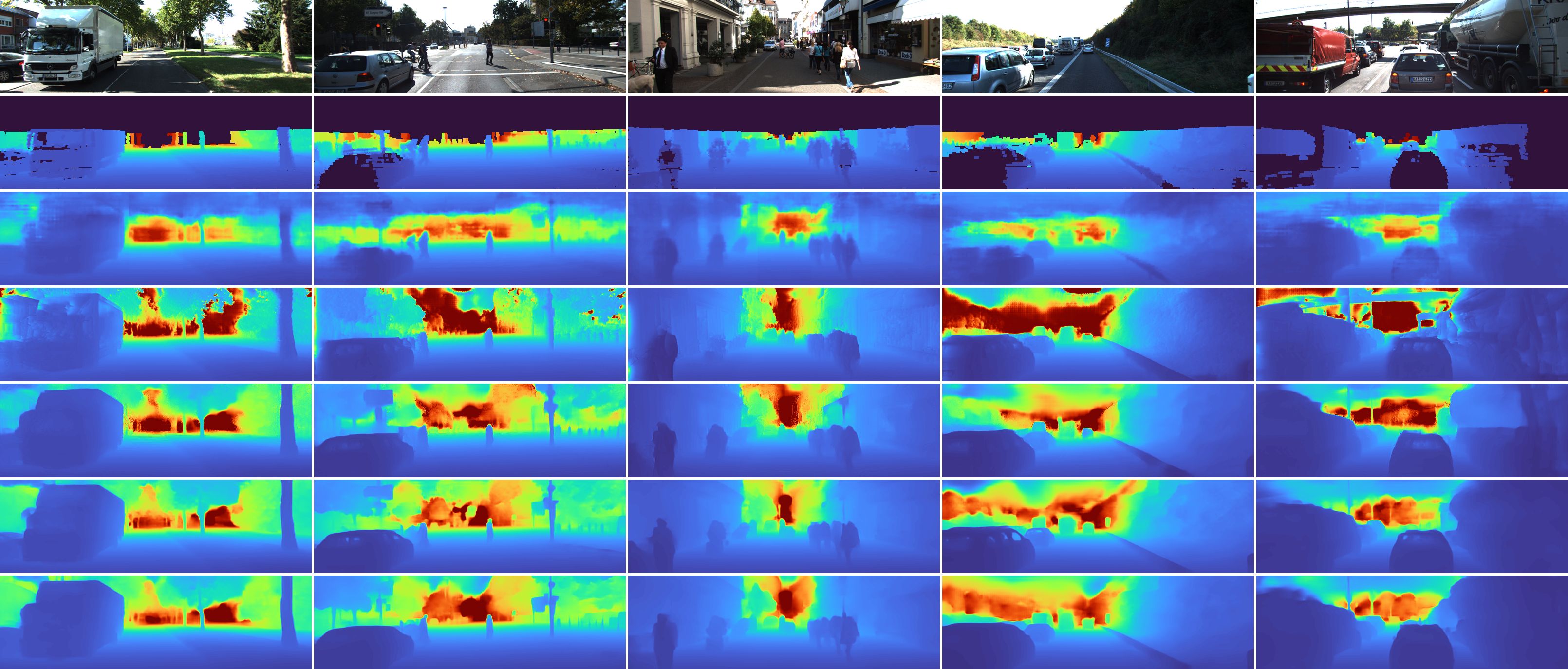}
    \caption{Qualitative results on the (KR) Eigen {\etal} testing slit \cite{Eigen:2014}. From top  to bottom: input images, their LiDAR-based depth GT (interpolated for visualization using \cite{Premebida:2014}), DORN, SharinGAN, PackNet-SfM, MonoDEVSNet VK\_v1 and VK\_v2.}
    \label{fig:kitti_eigen_qualitative}
\end{figure*}

\subsubsection{Qualitative results}
Figure \ref{fig:kitti_eigen_qualitative} presents qualitatively results relying on the depth color map commonly used in the MDE literature. We show results for representative methods in \Tab{SOTA_KITTI_eigen_absolute}, namely, DORN (LiDAR supervision), SharinGAN (stereo self-supervision and virtual-world supervision), PackNet-SfM (SfM self-supervision and ego-vehicle speed supervision), and MonoDEVSNet (Ours) using VK\_v1 and VK\_v2 (SfM self-supervision and virtual-world supervision). We also show the corresponding LiDAR-based GT. This GT shows that for LiDAR configurations such as the one used to acquire KITTI dataset, detecting some close vehicles may be problematic since only a few LiDAR points capture their presence. Despite being trained on LiDAR supervision, DORN provides more accurate depth information in these corner cases than the raw LiDAR, which is an example of the relevance of MDE in general. However, DORN shows worse results in these corner cases than the rest (SharinGAN/PackNet-SfM/Ours), even being more accurate in terms of MDE metrics, which focus on global assessment. SharinGAN has more difficulties than PackNet-SfM and our proposal for providing sharp borders in vertical objects/infra-structure ({\eg}, vehicles, pedestrians, traffic signs, trees). An interesting point to highlight is also the qualitative difference that we observe on our results depending on the use of VK version. In VK\_v1 data, vehicle windows appear as transparent to depth, like in many cases happens with LiDAR data, while in VK\_v2 they appear as solid. This is translated to the MDE results as we can observe comparing the two bottom rows of \Fig{kitti_eigen_qualitative}. Technically, we think the qualitative results of VK\_v2 make more sense since the windows are there at the given depth. However, what we would like to highlight is that we can select one option or another thanks to the use of virtual-world data. 

\begin{figure}
    \centering
    \includegraphics[width=\columnwidth]{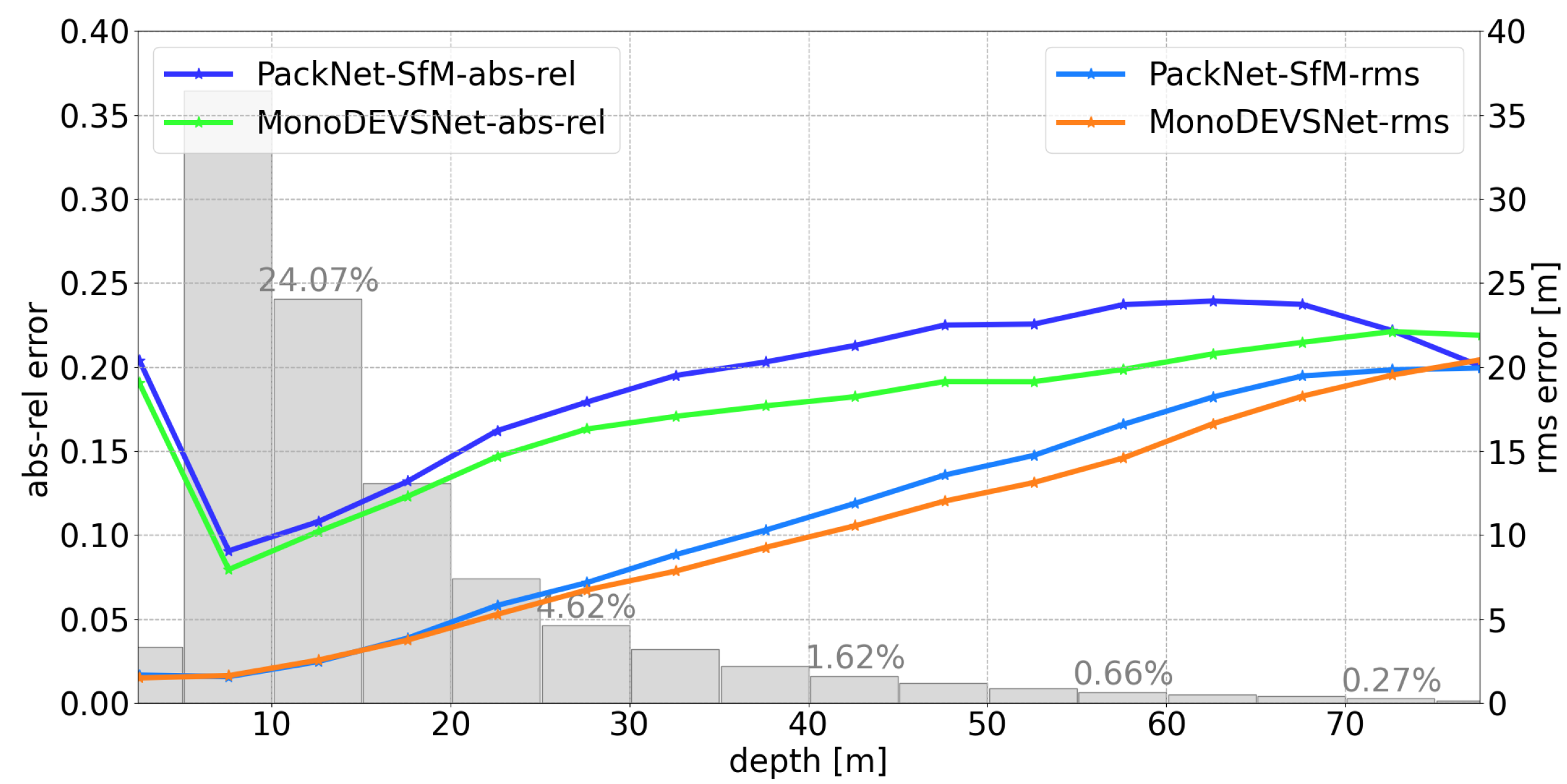}
    \caption{Abs-rel and rms errors as a function of depth, for KR Eigen testing split \cite{Eigen:2014}. The histogram of depth GT is shown with bars. We compare PackNet-SfM and MonoDEVSNet.}
    \label{fig:depth_error_as_a_function_of_depth}
\end{figure}

\begin{figure}
    \centering
    \includegraphics[width=\columnwidth]{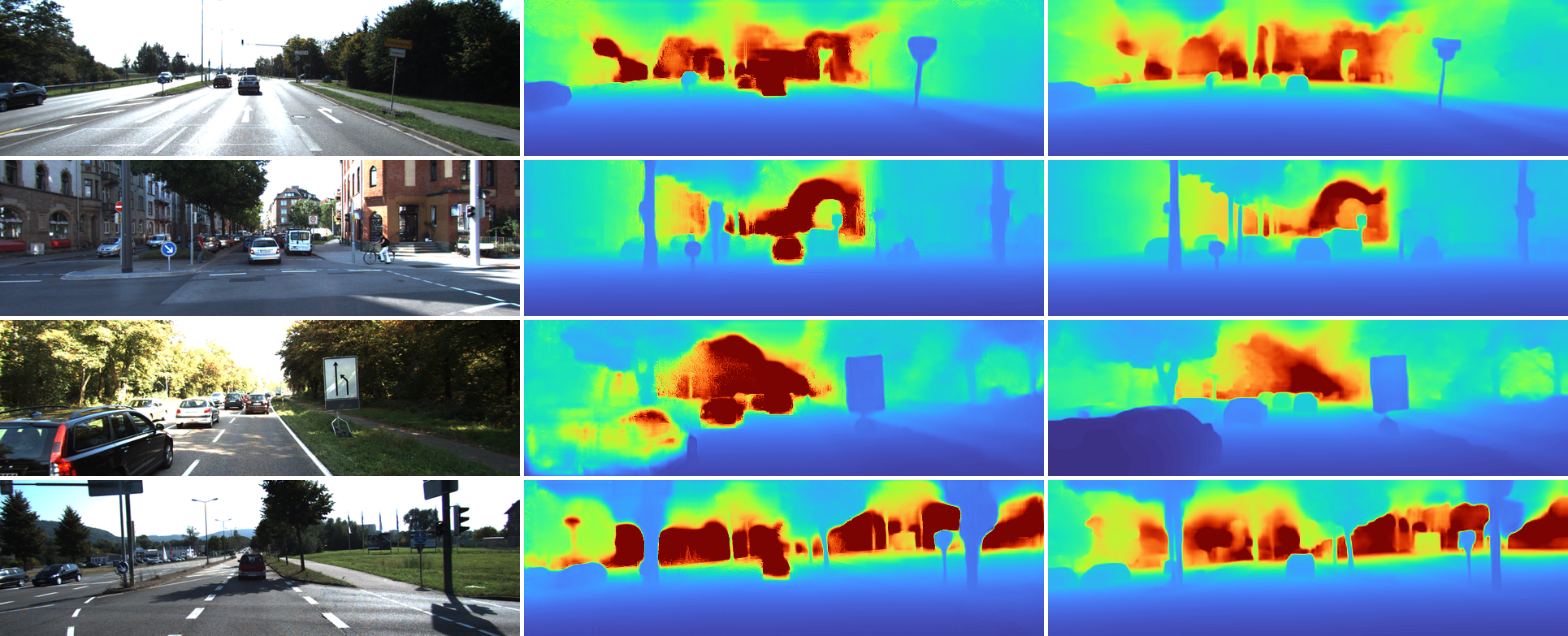}
    \caption{Qualitative results on KS data. From left to right: input images, PackNet-SfM, MonoDEVSNet.}
    \label{fig:kitti_2015_ghost_cars_qualitative}
\end{figure}

\begin{figure}
    \centering
    \includegraphics[width=\columnwidth]{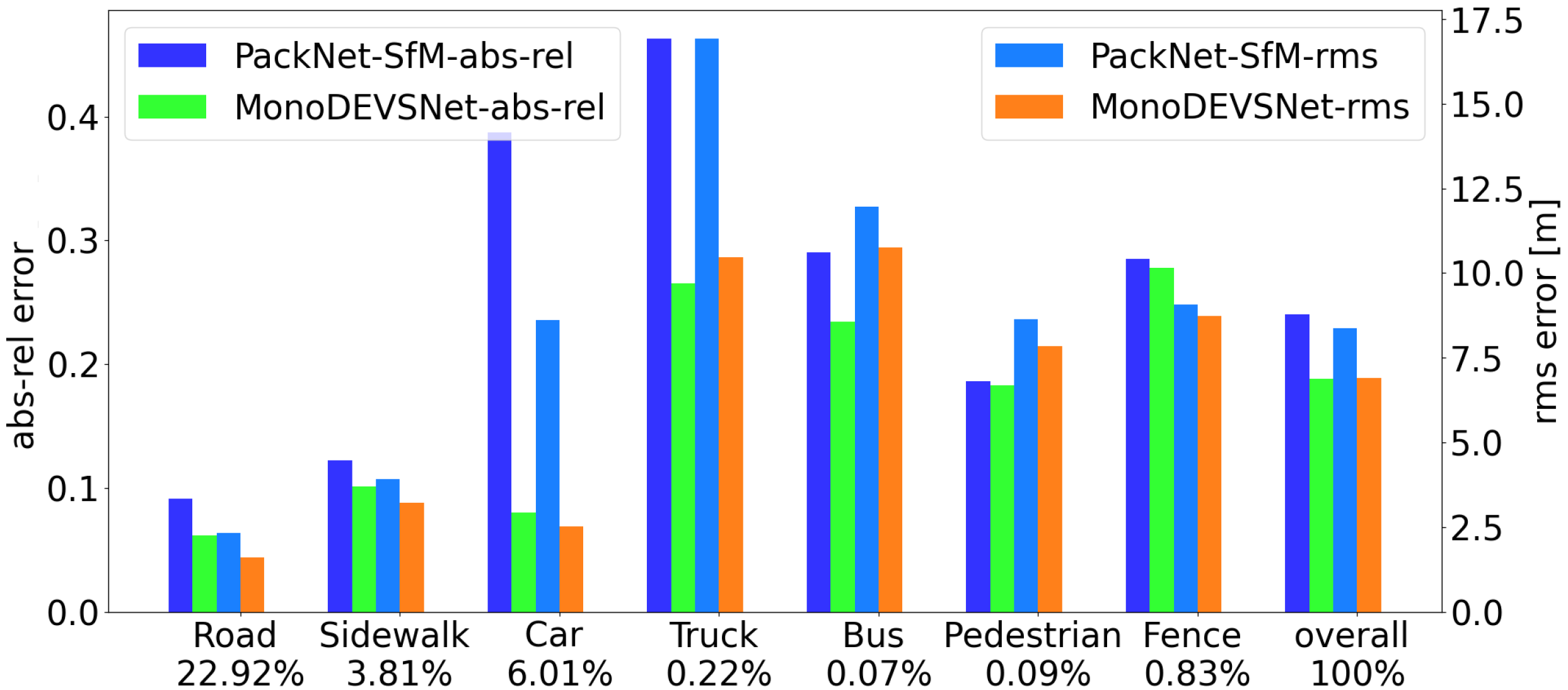}
    \caption{Per-class abs-rel and rms errors for KS, computed by averaging over the pixels of each class, for PackNet-SfM and MonoDEVSNet. The $\%$ of pixels of each class is shown.}
    \label{fig:per_class_analysis}
\end{figure}

\begin{figure}
    \centering
    \includegraphics[clip=true, trim=10 0 2 10, width=\columnwidth]{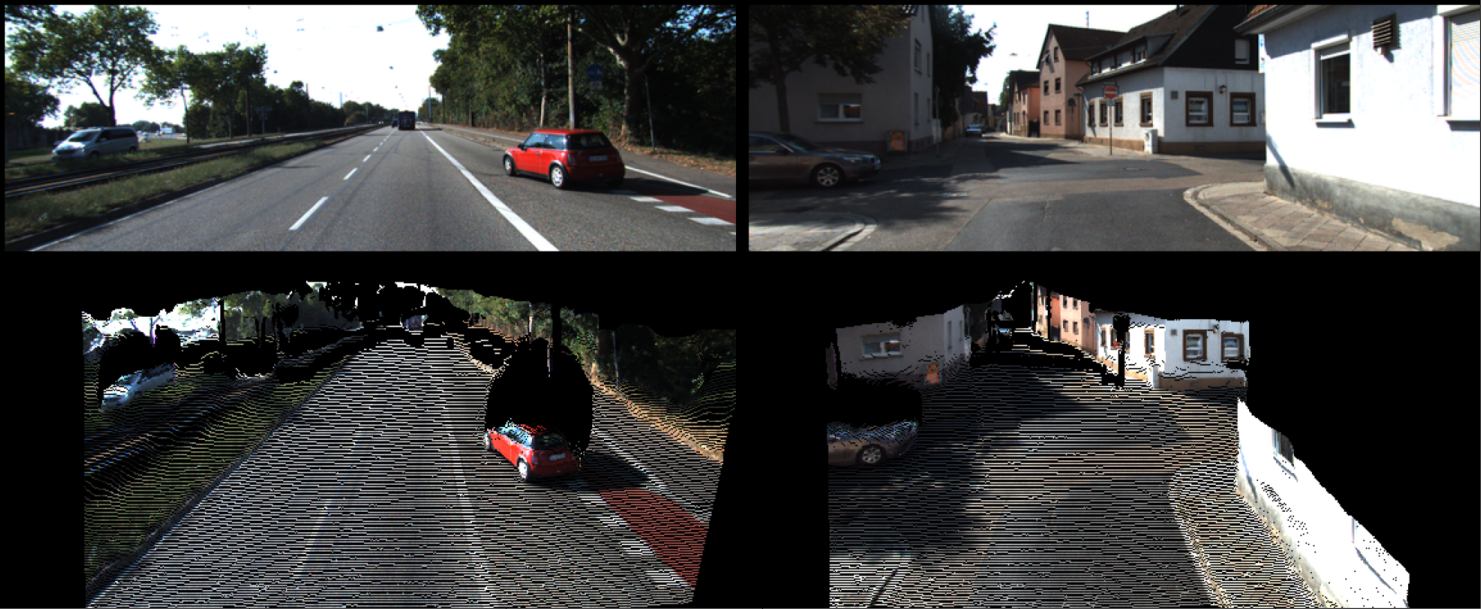}
    \caption{Point cloud representation on KR Eigen test split \cite{Eigen:2014} and KS data from left to right. From top to bottom: input images, MonoDEVSNet textured point cloud.}
    \label{fig:pcd_representation}
\end{figure}

\subsubsection{Additional insights}
In terms of qualitative results we think the best performing and most similar approaches are PackNet-SfM and MonoDEVSNet, both relying on real-world monocular systems. Thus, we perform a deeper comparison of them. First, following PackNet-SfM article \cite{Guizilini:20203D}, \Fig{depth_error_as_a_function_of_depth} plots the abs-rel error as a function of depth. Since this is a relative error, we also plot rms. Results are similar within a close range of up to $20$m. Within $20$m and $70$m, our proposal clearly outperforms PackNet-SfM and beyond $70$m both methods perform similarly. How these differences translate to abs-rel and rms global scores depends on the number of pixels falling in each distance range, which we show as an histogram in the same plot. We see how for the KR testing set most of the pixels fall in the $5-20$m depth range, where both methods perform more similarly. Second, we provide further comparative insights by using KS data since it has associated per-class semantic GT. Note that, although KS is a different data split than the one used in the experiments shown so far (KR), still is KITTI data; thus, we are not yet facing experiments about generalization. Figure \ref{fig:kitti_2015_ghost_cars_qualitative} compares qualitative results of PackNet-SfM {\vs} MonoDEVSNet. We can see how PackNet-SfM misses some vehicles that our proposal does not. We believe that these vehicles may be moving at a similar speed w.r.t the ego-vehicle, which may be problematic for pure SfM-based approaches and we hypothesize that virtual-world supervision can help to avoid this problem. Figure \ref{fig:per_class_analysis} shows the corresponding abs-rel metric per-class, focusing on the most relevant classes for driving. Note how the main differences between PackNet-SfM and MonoDEVSNet are observed on vehicles, especially on cars. 

Additional qualitative results are added in \Fig{pcd_representation}, where we can see how original images from KR and KS can be rendered as a textured point cloud. In particular, the viewpoint of these renders can change with respect to the original images thanks to the absolute depth values obtained with MonoDEVSNet.

\subsubsection{Generalization results}
\begin{figure}
    \centering
    \includegraphics[clip=true, trim=0 0 0 0, width=\columnwidth]{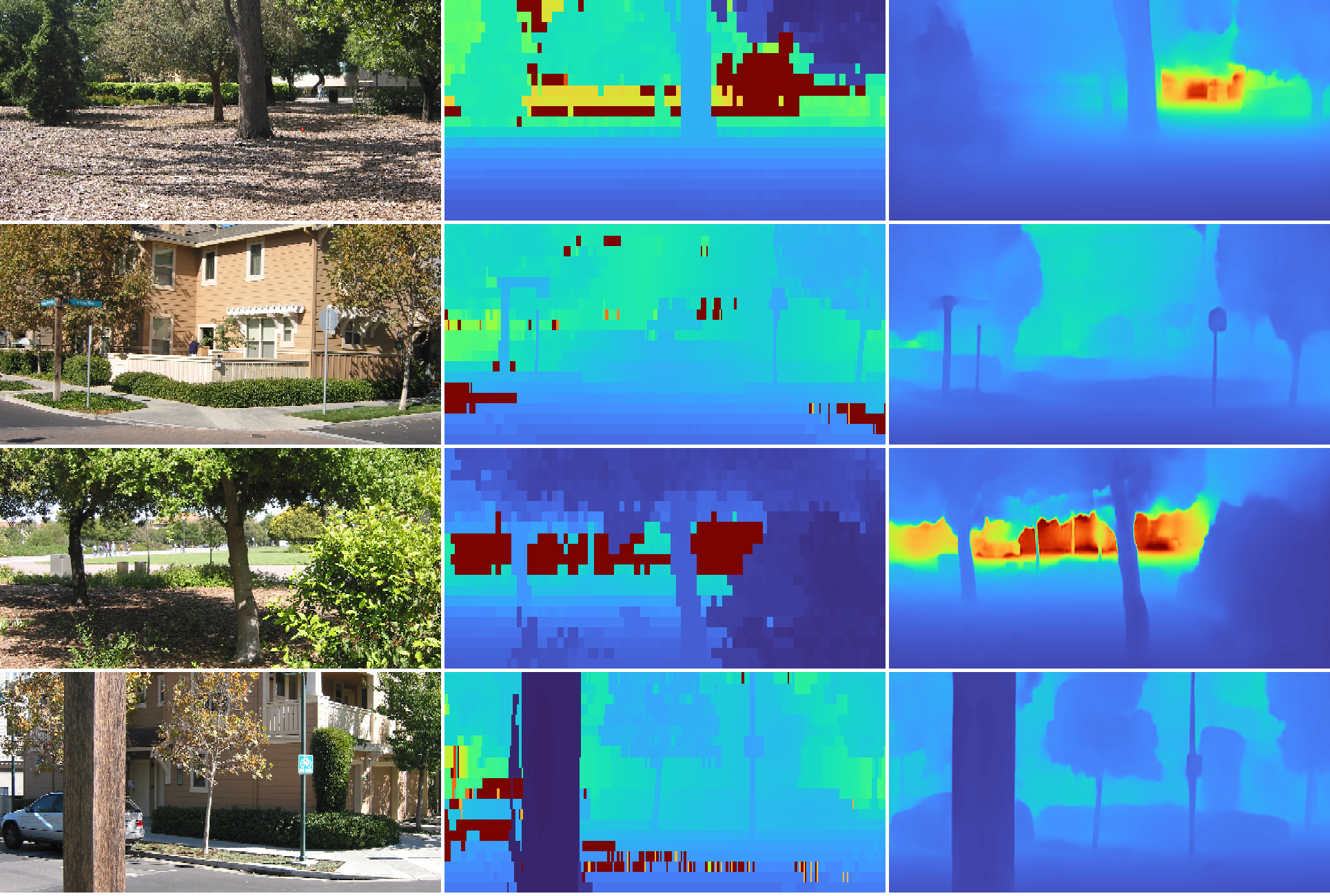}
    \caption{Qualitative results of MonoDEVSNet on Make3D. From left to right: input images, depth GT, MonoDEVSNet. }
    \label{fig:make3d_qualitative}
\end{figure}

\begin{figure}
    \centering
    \includegraphics[width=\columnwidth]{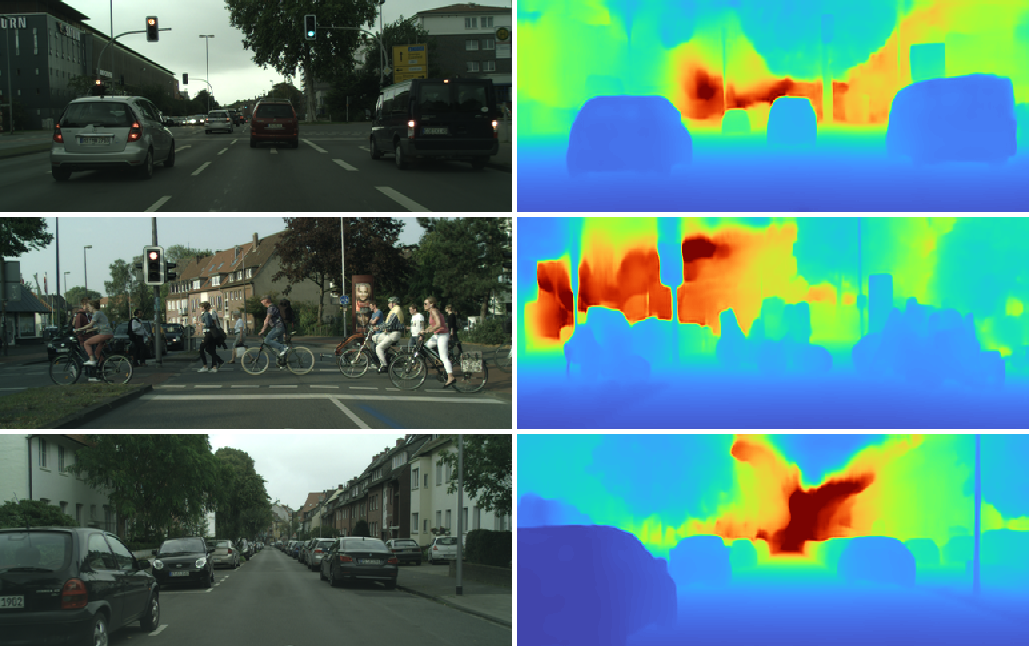}
    \caption{Qualitative results of MonoDEVSNet on Cityscapes dataset. From left to right: input images, MonoDEVSNet estimated depth maps.}
    \label{fig:cityscapes_qualitative}
\end{figure}

As done in the previous literature using VK to support MDE \cite{Kundu:2018AdaDepth, Zheng:2018T2Net, Zhao:2019GASDA, Pnvr:2020SharinGAN}, we assess generalization on Make3D dataset. As in this literature, we follow the standard data conditioning (cropping and resizing) for models trained on KR, as well as the standard protocol introduced in \cite{Godard:2017} to compute MDE evaluation metrics ({\eg} only depth below $70$m is considered). Table \ref{tab:sota_make3D} presents the quantitative results usually reported for Make3D, and ours. Note how, in generalization terms, our method also outperforms the rest. Moreover, \Fig{make3d_qualitative} shows how our proposal captures the depth structure even better than the depth GT, which is build from ${55\times305}$ depth maps acquired by a 3D scanner. In addition, we show qualitative results on Cityscapes dataset \cite{Cordts:2016} in \Fig{cityscapes_qualitative}. This dataset is captured in a real-world driving setup similar to car mounted videos of KITTI.

\begin{table}
\centering
\caption{\emph{\underline{Absolute depth}} results on Make3D testing set. All the shown methods use Make3D only for testing (generalization), except ($^1$) which fine-tunes on Make3D training set too.}
\label{tab:sota_make3D}
\begin{tabular}{|l|*{4}{c|}}\hline
Method &\makebox[3.5em]{abs-rel} &\makebox[3.5em]{sq-rel} &\makebox[3.5em]{rms} \\ \hline \hline
\cite{Zheng:2018T2Net} $T^2$Net  / VK\_v1             & 0.508      & 6.589      & 8.935 \\ \hline
\cite{Kundu:2018AdaDepth} AdaDepth-S$^1$ / VK\_v1 & 0.452      & 5.71       & 9.559 \\ \hline
\cite{Zhao:2019GASDA} GASDA / VK\_v1              & 0.403      & 6.709      & 10.424 \\ \hline
\cite{Pnvr:2020SharinGAN} SharinGAN / VK\_v1      & \IL{0.377} & 4.900      & 8.388 \\ \hline
MonoDEVSNet / VK\_v1                                     & 0.381      & \IL{3.997} & \B{7.949} \\ \hline 
MonoDEVSNet / VK\_v2                                     & \B 0.377   & \B 3.782   & \IL{8.011} \\ \hline
\end{tabular}
\end{table}

\subsubsection{Failure cases} 
\begin{figure}
    \centering
    \includegraphics[width=\columnwidth]{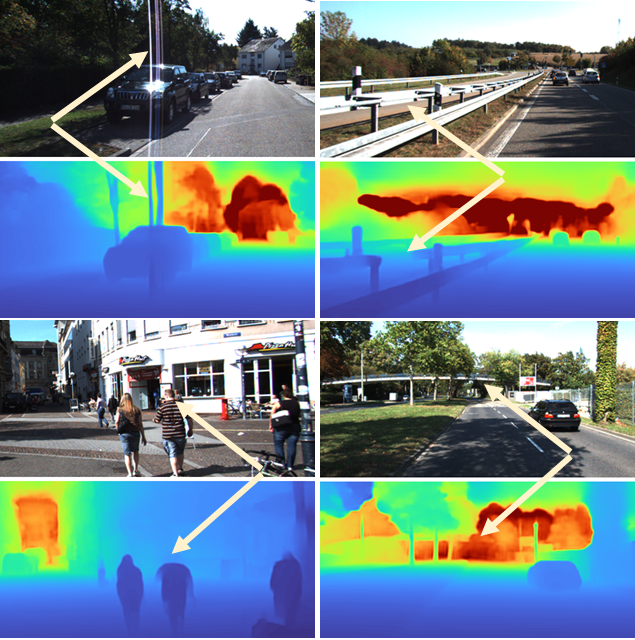}
    \caption{Failure cases in the depth map.}
    \label{fig:failurecases}
\end{figure}

\Fig{failurecases} shows qualitative results where some errors in the depth map are highlighted: (1) overexposed pixel columns lead to hallucinate a vertical structure; (2) saturated fence segments are not seen; (3) pedestrians are visible but with an approximate silhouette; (4) a bridge is not seen; (5) saturated skies do not appear as faraway. Thinking in the information required to drive and assuming that depth estimation is combined with semantic segmentation, we think that the cases (3) and (5) are not a problem and the (2) would be only if the unseen segment is too large (which is not the case in the shown example). The case (4) could be a problem for an autonomous bus/truck provided it does not fit below the bridge, but usually those would have predefined routes where this should not happen. However, (1) can be a problem depending where the hallucinated structure appears, {\eg}, in the example probably it would not be a problem, but it would be if the structure appears in the middle of the road. Behind some errors, we can find the lack of training data ({\eg}, for the bridge not seen). Behind others, we find extreme imaging conditions (like overexposed image areas). In the former case, we need to re-train by taking these cases into account; while, in the latter case, we need to prevent such undesired effects ({\eg} by using HDR camera settings). It is worth to mention that we have seen similar errors in other methods in the literature.

\subsubsection{Revisiting $\wmdeparenc$ architectures} We selected HRNet-W48 because it provides the most accurate results, however, we may need to sacrifice accuracy to reduce the computational burden. Thus, we have run more experiments with the final MonoDEVS training approach, just changing $\wmdeparenc$. These include representative architectures of ResNet and HRNet types, as well as, DenseNet \cite{Huang:2019DenseNet}. As ResNet, DenseNet does not need adding the pyramidal blocks ($\wmdeparpyr$).  Moreover, as we mentioned in \sSect{datasets}, to keep our experimental work manageable, we used 12K triplets (samples) from the real- and virtual-world training sets; however, it is possible to use $\sim\mbox{40K}$ samples from KR, which is the common practice in the literature (as those using KR in Tables \ref{tab:compare_netarch}-\ref{tab:SOTA_KITTI_eigen_absolute}). Likewise, we use $\sim\mbox{22K}$ samples from VK\_v2 as other literature methods in \Tab{sota_make3D} do with VK\_v1.  Table \ref{tab:all_architectures_and_39K} presents the corresponding results. The block $\sim\mbox{40K/22K}$ can be directly compared to the literature in  \Tab{SOTA_KITTI_eigen_absolute}). We see how, indeed, HRNet-W48 is the best in term of accuracy metrics. However, we see that DenseNet-121 offers the best trade-off between memory (MW) and computational (GFLOPS) requirements, offering real-time (FPS) with accuracy close to the state-of-the-art. If we need to reduce the computational burden and significantly increase the FPS, then ResNet-18 is a reasonable alternative.

\begin{table}
\centering
\caption{\underline{Absolute depth}. We provide final experimental results for different versions of three different architectures. In the upper block of the table  we have used the same training set as in previous experiments, {\ie}, 12K/12K from KR/VK\_v2. In the bottom block, as is usual in the literature, we use all the available training data, {\ie}, $\sim\mbox{40K}$ and $\sim\mbox{22K}$, respectively.}
\label{tab:all_architectures_and_39K}
\begin{tabular}{|c||*{10}{c|}}\hline
\makebox[5em]{$\wmdeparenc$ Backb.}
&\makebox[2em]{MW}&\makebox[4em]{GFLOPS}&\makebox[2.5em]{FPS} &\makebox[2.5em]{abs-rel}&\makebox[2.5em]{sq-rel}&\makebox[2em]{rms}&\makebox[2em]{$1.25$}&\makebox[2em]{$1.25^2$}\\\hline \hline
% 12K real and 12k synthetic images
ResNet-18    &   11.6   & \B 4.47   & \B 141.2  &   0.116       &   0.836       &   4.735       &   0.860       &   0.954       \\
%ResNet-50    &   25.6   &   10.14   &\IL{77.06} &   0.115       &   0.833       &   4.625       &   0.859       &   0.956      \\
ResNet-152   &   60.2   &   19.29   &   30.71   &   \IL{0.108}  &   \IL{0.759}  &   4.559       &   0.870       &   0.960       \\\hline
HRNet-W18    & \IL{9.5} &    8.29   &   15.79   &   0.109       &   0.773       &   4.524       &   \IL{0.871}  &   0.960       \\ 
HRNet-W48    &   65.3   &   40.04   &   15.48   &   \B 0.104    &   \B 0.721    &   \B 4.396    &   \B 0.880    &   \B 0.962    \\\hline
DenseNet-121 &  \B 6.9  &\IL{7.09}  &\IL{32.60} &   0.116       &   0.812       &   4.646       &   0.854       &   0.960       \\
DenseNet-161 &   26.5   &   19.21   &   24.87   &   0.111       &   0.763       &   \IL{4.516}  &   0.864       &   \IL{0.960}  \\\hline 
\Xhline{4\arrayrulewidth}
% ~40K real and ~22k synthetic images
ResNet-18    &   11.6   &  \B 4.47  & \B 141.2  &   0.114       &   0.838       &   4.734       &   0.860       &   0.954       \\
%ResNet-50    &   25.6   &   10.14   &\IL{77.06} &   0.111       &   0.814       &   4.595       &   0.869       &   0.959       \\
ResNet-152   &   60.2   &   19.29   &   30.71   &   \IL{0.104}  &   0.784       &   4.560       &   \IL{0.878}  &   0.0.960     \\\hline
HRNet-W18    & \IL{9.5} &    8.29   &   15.79   &   0.105       &   \IL{0.745}  &   4.470       &   0.877       &   0.961       \\ 
HRNet-W48    &   65.3   &   40.04   &   15.48   &   \B 0.101    &  \B 0.703     &   \B 4.413    &   \B 0.882    &   \B 0.962    \\\hline
%HRNet-W48-PP &   65.3   &   40.04   &   15.48   &   \B 0.100    &  \B 0.689     &   \B 4.317    &   \B 0.883    &   \B 0.962    \\\hline
DenseNet-121 & \B 6.9   & \IL{7.09} &\IL{32.60} &   0.111       &   0.786       &   4.536       &   0.870       &   0.960       \\
DenseNet-161 &   26.5   &   19.21   &   24.87   &   0.109       &   0.760       &   \IL{4.440}  &   0.873       &  \IL{0.962}   \\\hline 
\Xhline{4\arrayrulewidth}
\end{tabular}
\end{table}

\section{Conclusion}
\label{sec:conclusion}
For on-board perception, we have addressed monocular depth estimation by virtual-world supervision (MonoDEVS) and real-world SfM-inspired self-supervision; the former compensating for the inherent limitations of the latter. This challenging setting allows to rely on a monocular system not only at testing time, but also at training time; a cheap and scalable approach. We have designed a CNN, MonoDEVSNet, which seamlessly trains on real- and virtual-world data, exploiting semantic and depth supervision from the virtual-world data, and addressing the virtual-to-real domain gap by a relatively simple approach which does not add computational complexity in testing time. We have performed a comprehensive set of experiments assessing quantitative results in terms of relative and absolute depth, generalization, and we show the relevance of the components involved on MonoDEVSNet training. Our proposal yields state-of-the-art results within the SfM-based setting, even outperforming stereo-based self-supervised approaches. Qualitative results also confirm that MonoDEVSNet properly captures the depth structure of the images. As a result, we show the usefulness of leveraging virtual-world supervision to ultimately reach the upper-bound performance of methods based on LiDAR supervision. Therefore, our next steps will focus on analyzing the detailed differences between LiDAR-based supervision methods and MonoDEVSNet to find even better ways to benefit from virtual-world supervision.  

\appendix

\subsection{MonoDELSNet}
\label{appsec:monodelsnet}

\begin{figure}
    \centering
    \includegraphics[width=1.\linewidth]{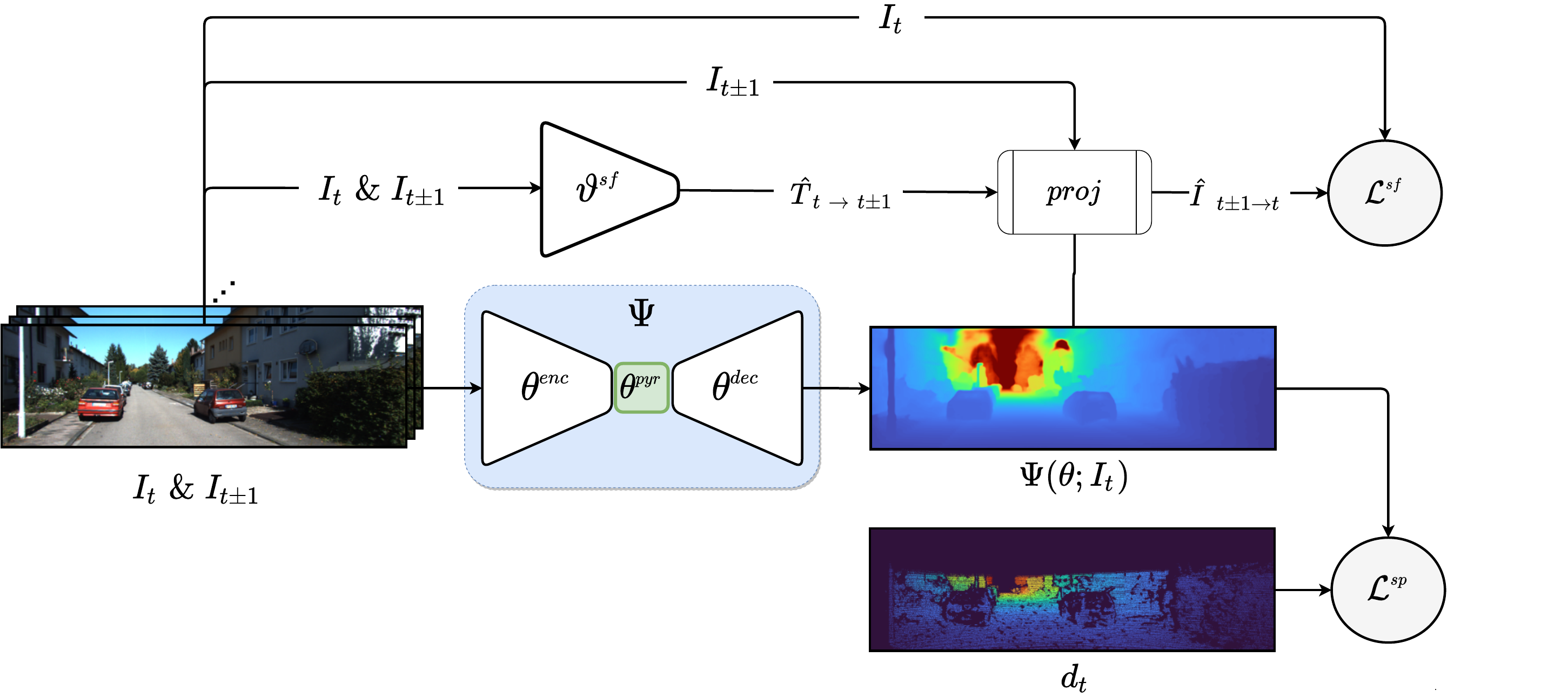}
    \caption{Training framework for MonoDELSNet, {\ie}, $\wmdepar{\inputimage}$ uses SfM self-supervision and LiDAR supervision. We show the involved images, GT from LiDAR, weights, and losses. }
    \label{fig:monodels_arch}
\end{figure}

We introduce \emph{Mono}cular Depth Estimation through \emph{L}iDAR \emph{S}upervision and SfM Self-Supervision (MonoDELSNet) for real-world datasets, which is an adaptation of the MonoDEVSNet proposal. In MonoDELSNet we replace the virtual-world supervision of MonoDEVSNet by real-world LiDAR supervision. In addition, as for these experiments supervision and SfM self-supervision come from the same domain, we have removed the domain classifier and gradient-reversal-layer (GRL) components present in MonoDEVSNet. Furthermore, as the supervision for semantic classes is not available in this case, we do not consider the class weighting mask used to train MonoDEVSNet. The modified proposal is summarized in \Fig{monodels_arch}.

\subsubsection{LiDAR Supervised loss: $\lsuplosspar{\wmde; {\RWSequences.\RWSequencesGT}}$}

Since we address an estimation problem and we use depth supervision (ground truth), captured with a LiDAR sensor, we use a corresponding loss, $\lsuploss$, based on the L1 metric. We denote the depth supervision as
$\RWSequencesGT=\{\dr{t}\}_{t=1}^{\Nr}$, 
where $\dr{t}$ is its depth map supervision for the corresponding frame in $\RWSequences=\{\ir{t}\}_{t=1}^{\Nr}$, and $\Nr$ is the number of frames with such supervision. Accordingly, we define the LiDAR-based loss function as:

\begin{equation}
\label{eq:sup-loss}
\lsuplosspar{\wmde; {\RWSequences.\RWSequencesGT}} = \sum_{t=1}^{\Nr} \LONEMETRIC{\wmdepar{\ir{t}}-\dr{t}} \enspace , 
\end{equation}

\noindent where $\wmdepar{\ir{t}}$ is the estimated depth for frame $\ir{t}$, being $\wmde=\{\wmdeparenc,\wmdeparpyr,\wmdepardec\}$. The SfM self-supervised loss is $\selflosspar{\wmde, \wauxself; {\RWSequences}}$, as defined in MonoDEVSNet. 

% \vspace{-0.3cm}

\subsubsection{Overall training procedure} 
\label{ssec:traininprocedure}
Algorithm \ref{alg:monodelsgradientcomputation} summarizes the steps to compute the needed gradient vectors for mini-batch optimization. In particular, the gradients related to MonoDELSNet' weights, $\wmde$, and those related to the auxiliary task (SfM self-supervision), {\ie}, $\wauxself$.

% \newpage
{
%\vspace{1cm}
\begin{algorithm}
\caption{Computing the gradients $\lossgradvalue{}{\wmde}$, $\lossgradvalue{}{\wauxself}$ for a mini-batch $\BRWSequences.\BRWSequencesGT\subset\RWSequences.\RWSequencesGT$. $\lossgrad{\xi}F(\xi)$ refers to back-propagation on $F(\xi)$ with respect to weights $\xi$. Analogous for $\lossgrad{\xi_i}F(\xi_1,\xi_2)$ regarding $F(\xi_1,\xi_2)$ and $\xi_i\subset\xi_1\cup\xi_2$.}
\label{alg:monodelsgradientcomputation}
\end{algorithm}
\begin{minipage}{\columnwidth}
{
\vspace{-0.3cm}
\centering Forward Passes with $\{\BRWSequences,\BRWSequencesGT\}$\par}
\vspace{-0.3cm}
\begin{align}
\ell^{\lsupervised}(\wmde)\gets&\lsuplosspar{\wmde;{\BRWSequences.\BRWSequencesGT}}\notag
\end{align}
{\centering Back-propagation for Supervision\par}
\vspace{-0.3cm}
\begin{align}
\lossgradvalue{\lsupervised}{\wmde}\gets&\lossgrad{\wmde}\ell^{\lsupervised}(\wmde)\notag
\end{align}
{\centering Forward Passes with $\BRWSequences$\par}
\vspace{-0.3cm}
\begin{align}
\ell^{\selfsupervised}(\wmde,\wauxself)\gets&\selflosspar{\wmde,\wauxself;\BRWSequences}\notag
\end{align}
{\centering Back-propagation for Self-supervision\par}
\vspace{-0.3cm}
\begin{align}
\lossgradvalue{\selfsupervised}{\wmde}\gets&\lossgrad{\wmde}\ell^{\selfsupervised}(\wmde,\wauxself)\notag
\end{align}
{\centering Setting the final gradient vectors\par}
\vspace{-0.3cm}
\begin{align}
\lossgradvalue{}{\wauxself}\gets&\lossgrad{\wauxself}\ell^{\selfsupervised}(\wmde,\wauxself)\notag\\[0.65em]
\selflosseq\gets&\ell^{\lsupervised}(\wmde)/\ell^{\selfsupervised}(\wmde,\wauxself)\notag\\[0.65em]
\lossgradvalue{}{\wmde}\gets&\lossgradvalue{\lsupervised}{\wmde} + \selflosseq\lossgradvalue{\selfsupervised}{\wmde}\notag
\end{align}
\end{minipage}
%\end{table}
}

\subsubsection{Experimental results}
%In this section, we explain about the experimental-setup and results. Firstly, the dataset is defined and training details are given. Then, we present and discuss our quantitative and qualitative results, comparing them with most recent supervision based MDE methods.
\paragraph{Dataset and Training details}
For training MonoDELSNet, we use the popular KITTI Raw (KR) dataset. Here, we follow Zhou~{\etal}~\cite{Zhou:2017} training-testing split. From the training split, we select ~40K monocular triplets, {\ie}, samples of the form $\{\ir{t-1}, \ir{t}, \ir{t+1}\}$, as well as an isolated sample of each triplet, $\{\ir{t}\}$, with densified LiDAR-based depth ground truth  $\{\dr{t}\}$. The former for providing SfM self-supervision, the latter for providing LiDAR supervision. For evaluation purposes, among the testing split (697 images) introduced by Eigen {\etal} \cite{Eigen:2014}, we considered the 652 isolated images with densified LiDAR-based depth supervision, {\ie}, as is common practice in the state-of-the-art methods on monocular depth estimation. HRNet-w48~\cite{Wang:2020HrNet} is used as backbone encoder, where its weights are initialized with ImageNet~\cite{Deng:2009} weights. Furthermore, the input images are processed at their original resolution, {\ie}, ${\sim1248\times384}$ pix.

\paragraph{Results and discussion}
We compare our method with LiDAR supervision methods such as VNL~\cite{Yin:2019} , DORN~\cite{Fu:2018DORN}, DPT~\cite{Ranftl:2021}, AdaBins~\cite{Bhat:2021} and with our combination of virtual-world supervision and SfM self-supervision, {\ie}, MonoDEVSNet. VNL combines supervision with 3D geometric constraints to improve the depth estimation accuracy. AdaBins proposes to learn adaptive bins per image. Similar to DORN, the idea is to discretize the depth range by dividing it into N bins and estimating the probability of bins per pixel. DPT uses a transformer-based backbone architecture to learn high-resolution representations.

\begin{figure}
    \centering
    \includegraphics[width=\linewidth]{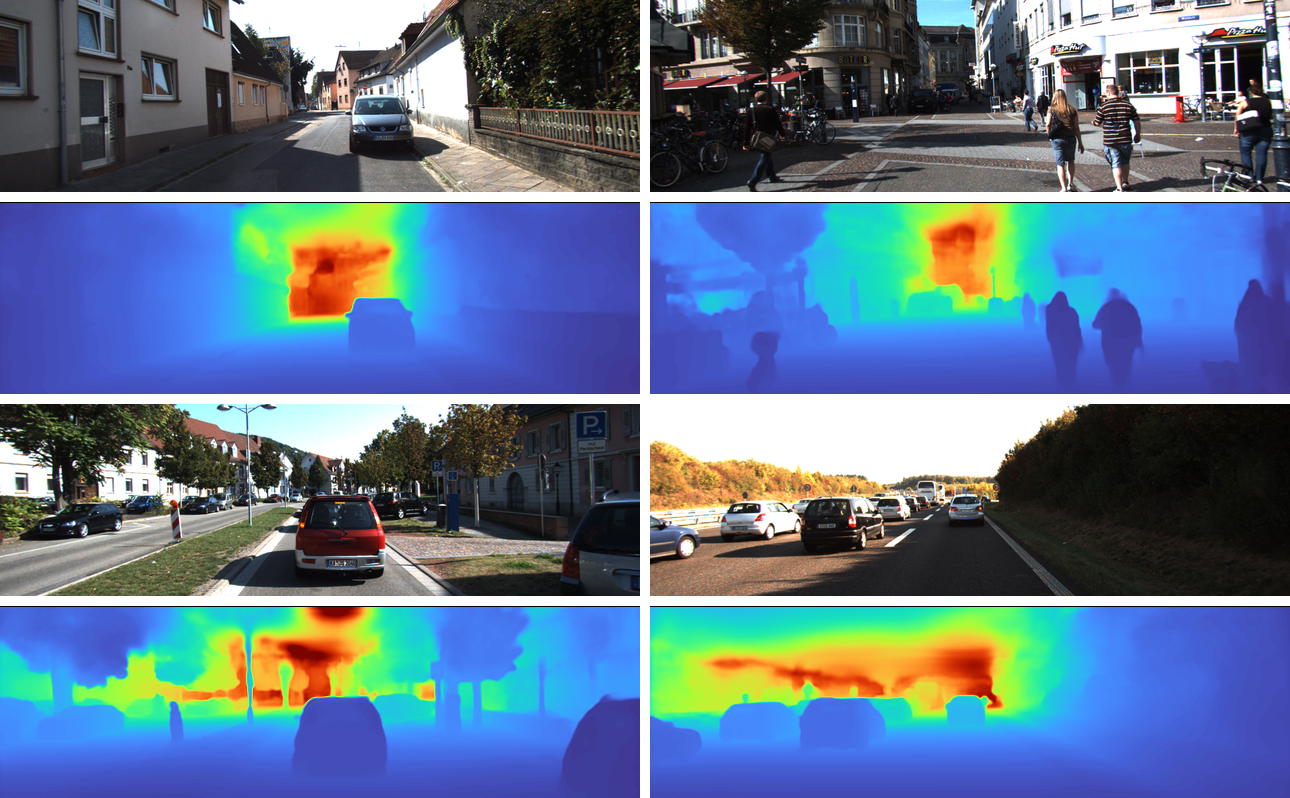}
    \caption{Qualitative results on the KR Eigen {\etal} testing split. From top to bottom, twice: input RGB image, and MonoDELSNet estimated depth maps.}
    \label{fig:qualitative}
\end{figure}

Again, we evaluate monocular depth estimation using the metrics proposed by Eigen {\etal} \cite{Eigen:2014}. \Tab{SOTA_KITTI_improved_absolute} shows our quantitative results compared to others of the state-of-the-art. Note how MonoDELSNet outperforms them. \Fig{qualitative} presents qualitatively results using a standard depth colormap. For completeness, we run experiments for MonoDEVSNet and MonoDELSNet based on the widely used resolution of $640\times192$ pix. Thanks to the LiDAR supervision, MonoDELSNet performs better. In addition, for the resolution of $1248\times384$ pix.,  we run MonoDELSNet with two different encoders, namely, ResNet-18 (also initialized from ImageNet for training) and HRNet-w48. The latter outperforms the former. We also assessed the results when the LiDAR supervision is used, but not the SfM-based self-supervision. We see that adding SfM-based self-supervision improves the results.

\begin{table}
    \centering
    \scriptsize
    \caption{\emph{\underline{Absolute depth}} results up to $80$m (see main text for details). ${(^1)}$ stands for models trained at the standard resolution $640\times192$ pix. ${(^2)}$ indicates the use of ResNet-18 instead of HRNet-w48 as encoder. ${(^3)}$ using LiDAR supervision only.}
    \label{tab:SOTA_KITTI_improved_absolute} 
    \begin{tabular}{|l||*{7}{c|}}\hline
        Method &\makebox[3em]{abs-rel}&\makebox[3em]{sq-rel}&\makebox[3em]{rms}&\makebox[3em]{rms-log}&\makebox[3em]{$1.25$}&\makebox[3em]{$1.25^2$}&\makebox[3em]{$ 1.25^3$}\\\hline \hline
        \cite{Yin:2019} VNL                 &   0.072   &   N/A     &   3.258   &   0.117   &   0.932   &   0.984   &   0.994 \\ \hline
        \cite{Fu:2018DORN} DORN             &   0.072   &   N/A     &   2.626   &   0.120   &   0.932   &   0.984   &   0.994  \\ \hline
        \cite{Ranftl:2021} DPT              &   0.062   &   0.222   &   2.573   &   0.092   &   0.959   &   0.995   &   0.999  \\ \hline
        \cite{Bhat:2021} AdaBins            &   0.058   &   0.190   &   2.360   &   0.088   &   0.964   &   0.995   &   0.999  \\ \hline
        %\Xhline{2\arrayrulewidth}
        MonoDEVSNet$^1$  &   0.073   &   0.298   &   2.802   &   0.108   &   0.939   &   0.990   &   0.988 \\ \hline
        %\Xhline{2\arrayrulewidth}
        MonoDELSNet$^1$                     &   0.065   &   0.281   &   2.951   &   0.102   &   0.948   &   0.992   &   0.998   \\ \hline
        MonoDELSNet$^2$                     &   0.062   &   0.217   &   2.502   &   0.097   &   0.954   &   0.994   &   0.998  \\ \hline
        MonoDELSNet     &   \B 0.053   &   \B 0.161   &   \B 2.101   &   \B 0.082   &   \B 0.969   &   \B 0.996   &   \B 0.999  \\ \hline
       % \Xhline{2\arrayrulewidth}
        MonoDELSNet$^3$                     &   0.057   &   0.217   &   2.613   &   0.092   &   0.960   &   0.994   &   0.998  \\ \hline
        %\Xhline{2\arrayrulewidth}
    \end{tabular}
\end{table}

\subsubsection{Conclusion} our MonoDEVNet framework can be adapted to work with LiDAR supervision and SfM self-supervision. This approach, here named as MonoDELSNet, reports state-of-the-art results on monocular depth estimation. 

\subsubsection{Publicly available code}
The MonoDELSNet is part of the MonoDEVSNet framework, thus, it can be found also here: \url{https://github.com/HMRC-AEL/MonoDEVSNet}

\ifCLASSOPTIONcaptionsoff
  \newpage
\fi



%\bibliographystyle{IEEEtran}

\bibliographystyle{IEEEtran}
%\bibliography{IEEEabrv}
\bibliography{main}

\begin{thebibliography}{10}
\providecommand{\url}[1]{#1}
\csname url@samestyle\endcsname
\providecommand{\newblock}{\relax}
\providecommand{\bibinfo}[2]{#2}
\providecommand{\BIBentrySTDinterwordspacing}{\spaceskip=0pt\relax}
\providecommand{\BIBentryALTinterwordstretchfactor}{4}
\providecommand{\BIBentryALTinterwordspacing}{\spaceskip=\fontdimen2\font plus
\BIBentryALTinterwordstretchfactor\fontdimen3\font minus
  \fontdimen4\font\relax}
\providecommand{\BIBforeignlanguage}[2]{{%
\expandafter\ifx\csname l@#1\endcsname\relax
\typeout{** WARNING: IEEEtran.bst: No hyphenation pattern has been}%
\typeout{** loaded for the language `#1'. Using the pattern for}%
\typeout{** the default language instead.}%
\else
\language=\csname l@#1\endcsname
\fi
#2}}
\providecommand{\BIBdecl}{\relax}
\BIBdecl

\bibitem{Dokhanchi:2021}
S.~Dokhanchi, B.~Mysore, K.~Mishra, and B.~Ottersten, ``Enhanced automotive
  target detection through {RADAR} and communications sensor fusion,'' in
  \emph{Int. Conf. on Acoustics, Speech, and Signal Processing (ICASSP)}, 2021.

\bibitem{Zhou:2018}
Y.~Zhou and O.~Tuzel, ``{VoxelNet}: End-to-end learning for point cloud based
  {3D} object detection,'' in \emph{Int. Conf. on Computer Vision and Pattern
  Recognition (CVPR)}, 2018.

\bibitem{Deac:2019}
S.~Deac, I.~Giosan, and S.~Nedevschi, ``Curb detection in urban traffic
  scenarios using lidars point cloud and semantically segmented color images,''
  in \emph{Intelligent Transportation Systems Conference (ITSC)}, 2019.

\bibitem{Cheng:2020}
X.~Cheng, Y.~Zhong, M.~Harandi, Y.~Dai, X.~Chang, T.~Drummond, H.~Li, and
  Z.~Ge, ``Hierarchical neural architecture search for deep stereo matching,''
  in \emph{Neural Information Processing Systems (NeurIPS)}, 2020.

\bibitem{Eigen:2014}
D.~Eigen, C.~Puhrsch, and R.~Fergus, ``Depth map prediction from a single image
  using a multi-scale deep network,'' in \emph{Neural Information Processing
  Systems (NeurIPS)}, 2014.

\bibitem{Liu:2016}
F.~Liu, C.~Shen, G.~Lin, and I.~Reid, ``Learning depth from single monocular
  images using deep convolutional neural fields,'' \emph{{IEEE} Trans. on
  Pattern Analysis and Machine Intelligence}, vol.~38, no.~10, pp. 2024--2039,
  2016.

\bibitem{Roy:2016}
A.~Roy and S.~Todorovic, ``Monocular depth estimation using neural regression
  forest,'' in \emph{Int. Conf. on Computer Vision and Pattern Recognition
  (CVPR)}, 2016.

\bibitem{Laina:2016}
I.~Laina, C.~Rupprecht, V.~Belagiannis, F.~Tombari, and N.~Navab, ``Deeper
  depth prediction with fully convolutional residual networks,'' in \emph{Int.
  Conf. on {3D} Vision {(3DV)}}, 2016.

\bibitem{Cao:2017}
Y.~Cao, Z.~Wu, and C.~Shen, ``Estimating depth from monocular images as
  classification using deep fully convolutional residual networks,''
  \emph{{IEEE} Trans. on Circuits and Systems for Video Technology}, 2017.

\bibitem{Fu:2018DORN}
H.~Fu, M.~Gong, C.~Wang, K.~Batmanghelich, and D.~Tao, ``Deep ordinal
  regression network for monocular depth estimation,'' in \emph{Int. Conf. on
  Computer Vision and Pattern Recognition (CVPR)}, 2018.

\bibitem{Gurram:2018}
A.~Gurram, O.~Urfalioglu, I.~Halfaoui, F.~Bouzaraa, and A.~L{\'o}pez,
  ``Monocular depth estimation by learning from heterogeneous datasets,'' in
  \emph{Intelligent Vehicles Symposium (IV)}, 2018.

\bibitem{He:2018}
L.~He, G.~Wang, and Z.~Hu, ``Learning depth from single images with deep neural
  network embedding focal length,'' \emph{{IEEE} Trans. on Image Processing},
  vol.~27, no.~9, pp. 4676--4689, 2018.

\bibitem{Xu:2018}
D.~Xu, W.~Wang, H.~Tang, H.~Liu, N.~Sebe, and E.~Ricci, ``Structured attention
  guided convolutional neural fields for monocular depth estimation,'' in
  \emph{Int. Conf. on Computer Vision and Pattern Recognition (CVPR)}, 2018.

\bibitem{Yin:2019}
W.~Yin, Y.~Liu, C.~Shen, and Y.~Yan, ``Enforcing geometric constraints of
  virtual normal for depth prediction,'' in \emph{International Conference on
  Computer Vision (ICCV)}, 2019.

\bibitem{Saxena:2007}
A.~Saxena, J.~Schulte, and A.~Ng, ``Depth estimation using monocular and stereo
  cues,'' in \emph{Int. Joint Conf. on Artificial Intelligence}, 2007.

\bibitem{Garg:2016}
R.~Garg, V.~Kumar, G.~Carneiro, and I.~Reid, ``Unsupervised cnn for single view
  depth estimation: Geometry to the rescue,'' in \emph{European Conference on
  Computer Vision (ECCV)}, 2016.

\bibitem{Godard:2017}
C.~Godard, O.~Aodha, and G.~Brostow, ``Unsupervised monocular depth estimation
  with left-right consistency,'' in \emph{Int. Conf. on Computer Vision and
  Pattern Recognition (CVPR)}, 2017.

\bibitem{Pillai:2019}
S.~Pillai, R.~Ambru{\c{s}}, and A.~Gaidon, ``{SuperDepth}: Self-supervised,
  super-resolved monocular depth estimation,'' in \emph{Int. Conf. on Robotics
  and Automation (ICRA)}, 2019.

\bibitem{Zhou:2017}
T.~Zhou, M.~Brown, N.~Snavely, and D.~Lowe, ``Unsupervised learning of depth
  and ego-motion from video,'' in \emph{Int. Conf. on Computer Vision and
  Pattern Recognition (CVPR)}, 2017.

\bibitem{Yin:2018GeoNet}
Z.~Yin and J.~Shi, ``{GeoNet}: Unsupervised learning of dense depth, optical
  flow and camera pose,'' in \emph{Int. Conf. on Computer Vision and Pattern
  Recognition (CVPR)}, 2018.

\bibitem{Zhao:2020}
W.~Zhao, S.~Liu, Y.~Shu, and Y.-J. Liu, ``Towards better generalization: Joint
  depth-pose learning without {PoseNet},'' in \emph{Int. Conf. on Computer
  Vision and Pattern Recognition (CVPR)}, 2020.

\bibitem{Guizilini:20203D}
V.~Guizilini, R.~Ambrus, S.~Pillai, A.~Raventos, and A.~Gaidon, ``{3D} packing
  for self-supervised monocular depth estimation,'' in \emph{Int. Conf. on
  Computer Vision and Pattern Recognition (CVPR)}, 2020.

\bibitem{Godard:2019MonoDepth2}
C.~Godard, O.~Mac~Aodha, M.~Firman, and G.~J. Brostow, ``Digging into
  self-supervised monocular depth estimation,'' in \emph{International
  Conference on Computer Vision (ICCV)}, 2019.

\bibitem{Kuznietsov:2017}
Y.~Kuznietsov, J.~St{\"u}ckler, and B.~Leibe, ``Semi-supervised deep learning
  for monocular depth map prediction,'' in \emph{Int. Conf. on Computer Vision
  and Pattern Recognition (CVPR)}, 2017.

\bibitem{He:2018wearable}
L.~He, C.~Chen, T.~Zhang, H.~Zhu, and S.~Wan, ``Wearable depth camera:
  Monocular depth estimation via sparse optimization under weak supervision,''
  \emph{{IEEE} Accesss}, vol.~6, pp. 41\,337--41\,345, 2018.

\bibitem{Guizilini:2020}
V.~Guizilini, J.~Li, R.~Ambrus, S.~Pillai, and A.~Gaidon, ``Robust
  semi-supervised monocular depth estimation with reprojected distances,'' in
  \emph{Conference on Robot Learning (CoRL)}, 2020.

\bibitem{De:2021}
R.~{de Queiroz}~Mendes, E.~G. Ribeiro, N.~{dos Santos}~Rosa, and V.~Grassi~Jr,
  ``On deep learning techniques to boost monocular depth estimation for
  autonomous navigation,'' \emph{Robotics and Autonomous Systems}, vol. 136, p.
  103701, February 2021.

\bibitem{Gaidon:2016}
A.~Gaidon, Q.~Wang, Y.~Cabon, and E.~Vig, ``Virtual worlds as proxy for
  multi-object tracking analysis,'' in \emph{Int. Conf. on Computer Vision and
  Pattern Recognition (CVPR)}, 2016.

\bibitem{Cabon:2020}
Y.~Cabon, N.~Murray, and M.~Humenberger, ``Virtual {KITTI} 2,''
  arXiv:2001.10773, 2020.

\bibitem{Ros:2016}
G.~Ros, L.~Sellart, J.~Materzyska, D.~V\'azquez, and A.~L\'opez, ``The
  {SYNTHIA} dataset: a large collection of synthetic images for semantic
  segmentation of urban scenes,'' in \emph{Int. Conf. on Computer Vision and
  Pattern Recognition (CVPR)}, 2016.

\bibitem{Mayer:2016}
N.~Mayer, E.~Ilg, P.~Hausser, P.~Fischer, D.~Cremers, A.~Dosovitskiy, and
  T.~Brox, ``A large dataset to train convolutional networks for disparity,
  optical flow, and scene flow estimation,'' in \emph{Int. Conf. on Computer
  Vision and Pattern Recognition (CVPR)}, 2016.

\bibitem{Richter:2017}
S.~R. Richter, Z.~Hayder, and V.~Koltun, ``Playing for benchmarks,'' in
  \emph{International Conference on Computer Vision (ICCV)}, 2017.

\bibitem{Shah:2017}
S.~Shah, D.~Dey, C.~Lovett, and A.~Kapoor, ``{AirSim}: High-fidelity visual and
  physical simulation for autonomous vehicles,'' in \emph{Int. Conf. on Field
  and Service Robotics (FSR)}, 2017.

\bibitem{Dosovitskiy:2017}
A.~Dosovitskiy, G.~Ros, F.~Codevilla, A.~L\'opez, and V.~Koltun, ``{CARLA}: An
  open urban driving simulator,'' in \emph{Conference on Robot Learning
  (CoRL)}, 2017.

\bibitem{Zheng:2018T2Net}
C.~Zheng, T.-J. Cham, and J.~Cai, ``{T2Net}: Synthetic-to-realistic translation
  for solving single-image depth estimation tasks,'' in \emph{European
  Conference on Computer Vision (ECCV)}, 2018.

\bibitem{Kundu:2018AdaDepth}
J.~Nath~Kundu, P.~Krishna~Uppala, A.~Pahuja, and R.~Venkatesh~Babu,
  ``{AdaDepth}: Unsupervised content congruent adaptation for depth
  estimation,'' in \emph{Int. Conf. on Computer Vision and Pattern Recognition
  (CVPR)}, 2018.

\bibitem{Zhao:2019GASDA}
S.~Zhao, H.~Fu, M.~Gong, and D.~Tao, ``Geometry-aware symmetric domain
  adaptation for monocular depth estimation,'' in \emph{Int. Conf. on Computer
  Vision and Pattern Recognition (CVPR)}, 2019.

\bibitem{Pnvr:2020SharinGAN}
K.~PNVR, H.~Zhou, and D.~Jacobs, ``{SharinGAN}: Combining synthetic and real
  data for unsupervised geometry estimation,'' in \emph{Int. Conf. on Computer
  Vision and Pattern Recognition (CVPR)}, 2020.

\bibitem{Cheng:2020S3Net}
B.~Cheng, I.~S. Saggu, R.~Shah, G.~Bansal, and D.~Bharadia, ``{S3Net}:
  Semantic-aware self-supervised depth estimation with monocular videos and
  synthetic data,'' in \emph{European Conference on Computer Vision (ECCV)},
  2020.

\bibitem{Csurka:2017}
G.~Csurka, \emph{A Comprehensive Survey on Domain Adaptation for Visual
  Applications}, ser. Advances in Computer Vision and Pattern
  Recognition.\hskip 1em plus 0.5em minus 0.4em\relax Springer, 2017, ch.~1.

\bibitem{Wang-Deng:2018}
M.~Wang and W.~Deng, ``Deep visual domain adaptation: {A} survey,''
  \emph{Neurocomputing}, vol. 312, pp. 135--153, October 2018.

\bibitem{Wilson:2020}
G.~Wilson and D.~Cook, ``A survey of unsupervised deep domain adaptation,''
  \emph{ACM Transactions on Intelligent Systems and Technology}, vol.~11,
  no.~5, 2020.

\bibitem{Ganin:2015}
Y.~Ganin and V.~Lempitsky, ``Unsupervised domain adaptation by
  backpropagation,'' in \emph{Int. Conf. on Machine Learning (ICML)}, 2015.

\bibitem{Ganin:2016}
Y.~Ganin, E.~Ustinova, H.~Ajakan, P.~Germain, H.~Larochelle, F.~Laviolette,
  M.~Marchand, and V.~Lempitsky, ``Domain-adversarial training of neural
  networks,'' \emph{The Journal of Machine Learning Research}, vol.~17, no.~1,
  pp. 2096--2030, 2016.

\bibitem{Liu:2010}
B.~Liu, S.~Gould, and D.~Koller, ``Single image depth estimation from predicted
  semantic labels,'' in \emph{Int. Conf. on Computer Vision and Pattern
  Recognition (CVPR)}, 2010.

\bibitem{Ladicky:2014}
L.~Ladicky, J.~Shi, and M.~Pollefeys, ``Pulling things out of perspective,'' in
  \emph{Int. Conf. on Computer Vision and Pattern Recognition (CVPR)}, 2014.

\bibitem{Srikakulapu:2015}
V.~{Srikakulapu}, H.~{Kumar}, S.~{Gupta}, and K.~S. {Venkatesh}, ``Depth
  estimation from single image using defocus and texture cues,'' in
  \emph{National Conference on Computer Vision, Pattern Recognition, Image
  Processing and Graphics (NCVPRIPG)}, 2015.

\bibitem{Mousavian:2016}
A.~Mousavian, H.~Pirsiavash, and J.~Ko{\v{s}}eck{\'a}, ``Joint semantic
  segmentation and depth estimation with deep convolutional networks,'' in
  \emph{Int. Conf. on {3D} Vision {(3DV)}}, 2016.

\bibitem{Jafari:2017}
O.~Jafari, O.~Groth, A.~Kirillov, M.~Yang, and C.~Rother, ``Analyzing modular
  {CNN} architectures for joint depth prediction and semantic segmentation,''
  in \emph{Int. Conf. on Robotics and Automation (ICRA)}, 2017.

\bibitem{Jiao:2018}
J.~Jiao, Y.~Cao, Y.~Song, and R.~Lau, ``Look deeper into depth: Monocular depth
  estimation with semantic booster and attention-driven loss,'' in
  \emph{European Conference on Computer Vision (ECCV)}, 2018.

\bibitem{Guizilini:2020semantic}
V.~Guizilini, R.~Hou, J.~Li, R.~Ambrus, and A.~Gaidon, ``Semantically-guided
  representation learning for self-supervised monocular depth,'' in \emph{Int.
  Conf. on Learning Representation (ICLR)}, 2020.

\bibitem{Chen:2019}
P.-Y. Chen, A.~H. Liu, Y.-C. Liu, and Y.-C.~F. Wang, ``Towards scene
  understanding: Unsupervised monocular depth estimation with semantic-aware
  representation,'' in \emph{Int. Conf. on Computer Vision and Pattern
  Recognition (CVPR)}, 2019.

\bibitem{Ozyesil:2017}
O.~Özyeşil, V.~Voroninski, R.~Basri, and A.~Singer, ``A survey of structure
  from motion,'' \emph{Acta Numerica}, vol.~26, pp. 305--364, 1st May 2017.

\bibitem{Wang:2004}
Z.~Wang, A.~C. Bovik, H.~R. Sheikh, and E.~P. Simoncelli, ``Image quality
  assessment: from error visibility to structural similarity,'' \emph{{IEEE}
  Trans. on Image Processing}, vol.~13, no.~4, pp. 600--612, 2004.

\bibitem{Goodfellow:2014}
I.~Goodfellow, J.~Pouget-Abadie, M.~Mirza, B.~Xu, D.~Warde-Farley, S.~Ozair,
  A.~Courville, and Y.~Bengio, ``Generative adversarial nets,'' in \emph{Neural
  Information Processing Systems (NeurIPS)}, 2014.

\bibitem{Choi:2020}
Y.~Choi, Y.~Uh, J.~Yoo, and J.-W. Ha, ``{StarGAN v2}: Diverse image synthesis
  for multiple domains,'' in \emph{Int. Conf. on Computer Vision and Pattern
  Recognition (CVPR)}, 2020.

\bibitem{Zhu:2017}
J.-Y. Zhu, T.~Park, P.~Isola, and A.~A. Efros, ``Unpaired image-to-image
  translation using cycle-consistent adversarial networks,'' in
  \emph{International Conference on Computer Vision (ICCV)}, 2017.

\bibitem{Chen:2018GradNorm}
Z.~Chen, V.~Badrinarayanan, C.-Y. Lee, and A.~Rabinovich, ``{GradNorm}:
  Gradient normalization for adaptive loss balancing in deep multitask
  networks,'' in \emph{Int. Conf. on Machine Learning (ICML)}, 2018.

\bibitem{Geiger:2013}
A.~Geiger, P.~Lenz, C.~Stiller, and R.~Urtasun, ``Vision meets robotics: The
  {KITTI} dataset,'' \emph{International Journal of Robotics Research},
  vol.~32, no.~11, pp. 1231--1237, 2013.

\bibitem{Menze:2015}
M.~Menze and A.~Geiger, ``Object scene flow for autonomous vehicle,'' in
  \emph{Int. Conf. on Computer Vision and Pattern Recognition (CVPR)}, 2015.

\bibitem{Saxena:2009}
A.~Saxena, M.~Sun, and A.~Ng., ``{Make3D}: Learning {3D} scene structure from a
  single still image,'' \emph{{IEEE} Trans. on Pattern Analysis and Machine
  Intelligence}, vol.~31, no.~5, pp. 824--840, 2009.

\bibitem{Wang:2020HrNet}
J.~Wang, K.~Sun, T.~Cheng, B.~Jiang, C.~Deng, Y.~Zhao, D.~Liu, Y.~Mu, M.~Tan,
  X.~Wang, W.~Liu, and B.~Xiao, ``Deep high-resolution representation learning
  for visual recognition,'' \emph{{IEEE} Trans. on Pattern Analysis and Machine
  Intelligence}, April 2020.

\bibitem{Paszke:2019pytorch}
A.~Paszke, S.~Gross, F.~Massa, A.~Lerer, J.~Bradbury, G.~Chanan, T.~Killeen,
  Z.~Lin, N.~Gimelshein, L.~Antiga, A.~Desmaison, A.~Kopf, E.~Yang, Z.~DeVito,
  M.~Raison, A.~Tejani, S.~Chilamkurthy, B.~Steiner, L.~Fang, J.~Bai, and
  S.~Chintala, ``{PyTorch}: An imperative style, high-performance deep learning
  library,'' in \emph{Neural Information Processing Systems (NeurIPS)}, 2019.

\bibitem{Kingma:2015}
D.~Kingma and J.~Ba, ``Adam : A method for stochastic optimization,'' in
  \emph{Int. Conf. on Learning Representation (ICLR)}, 2015.

\bibitem{Deng:2009}
J.~Deng, W.~Dong, R.~Socher, L.-J. Li, K.~Li, and L.~Fei-Fei, ``{ImageNet}: A
  large-scale hierarchical image database,'' in \emph{Int. Conf. on Computer
  Vision and Pattern Recognition (CVPR)}, 2009.

\bibitem{Cordts:2016}
M.~Cordts, M.~Omran, S.~Ramos, T.~Rehfeld, M.~Enzweiler, R.~Benenson,
  U.~Franke, S.~Roth, and B.~Schiele, ``The {Cityscapes} dataset for semantic
  urban scene understanding,'' in \emph{Int. Conf. on Computer Vision and
  Pattern Recognition (CVPR)}, 2016.

\bibitem{Premebida:2014}
C.~Premebida, J.~Carreira, J.~Batista, and U.~Nunes, ``Pedestrian detection
  combining {RGB} and dense {LiDAR} data,'' in \emph{Int. Conf. on Intelligent
  Robots and Systems (IROS)}, 2014.

\bibitem{Huang:2019DenseNet}
G.~Huang, Z.~Liu, G.~Pleiss, L.~Van Der~Maaten, and K.~Weinberger,
  ``Convolutional networks with dense connectivity,'' \emph{{IEEE} Trans. on
  Pattern Analysis and Machine Intelligence}, May 2019.

\bibitem{Ranftl:2021}
R.~Ranftl, A.~Bochkovskiy, and V.~Koltun, ``Vision transformers for dense
  prediction,'' in \emph{Int. Conf. on Computer Vision and Pattern Recognition
  (CVPR)}, 2021.

\bibitem{Bhat:2021}
S.~F. Bhat, I.~Alhashim, and P.~Wonka, ``Adabins: Depth estimation using
  adaptive bins,'' in \emph{Int. Conf. on Computer Vision and Pattern
  Recognition (CVPR)}, 2021.

\end{thebibliography}



%\vfill



\end{document}